\documentclass[10pt,twocolumn,letterpaper]{article}

\usepackage{cvpr}              

\usepackage[dvipsnames]{xcolor}

\definecolor{cvprblue}{rgb}{0.21,0.49,0.74}
\usepackage[pagebackref,breaklinks,colorlinks,citecolor=cvprblue]{hyperref}
\usepackage{bbding}
\usepackage{graphicx}
\usepackage{multirow}
\usepackage{multicol}
\usepackage{colortbl}
\usepackage{tabularx}
\usepackage{indentfirst}
\usepackage[accsupp]{axessibility}

\def\paperID{8236} 
\def\confName{CVPR}
\def\confYear{2025}

\title{Agri-LLaVA: Knowledge-Infused Large Multimodal Assistant on Agricultural Pests and Diseases}

\author{
Liqiong Wang$^{1,2\dagger}$,
Teng Jin$^{1,2\dagger}$,
Jinyu Yang$^{3\dagger}$, 
Ale\v{s} Leonardis$^{4}$,
Fangyi Wang$^{1,2}$,
Feng Zheng$^{5*}$ \\
$^1$Hubei Key Laboratory of Intelligent Vision Based Monitoring for \\Hydroelectric Engineering, China Three Gorges University \\
$^2$College of Computer and Information Technology, China Three Gorges University \\
$^3$Tapall.ai \
$^4$University of Birmingham \
$^5$Southern University of Science and Technology \\ 
{\tt\small
\{liqiong.wang11,Tengj0209,jinyu.yang96\}@outlook.com}\\
\tt\small
a.leonardis@cs.bham.ac.uk\
fy\_wang@ctgu.edu.cn\ 
f.zheng@ieee.org}

\begin{document}
\maketitle

\let\thefootnote\relax\footnotetext{$\dagger$ Equal contribution. $*$ Corresponding author.}
\let\thefootnote\relax\footnotetext{This work was done during Liqiong Wang visited SUSTech VIP lab.}

\begin{abstract}
\vspace{-0.7em}
In the general domain, large multimodal models (LMMs) have achieved significant advancements, yet challenges persist in applying them to specific fields, especially agriculture.
As the backbone of the global economy, agriculture confronts numerous challenges, with pests and diseases being particularly concerning due to their complexity, variability, rapid spread, and high resistance. 
This paper specifically addresses these issues.
We construct the first multimodal instruction-following dataset in the agricultural domain, covering over 221 types of pests and diseases with approximately 400,000 data entries. 
This dataset aims to explore and address the unique challenges in pest and disease control. 
Based on this dataset, we propose a knowledge-infused training method to develop Agri-LLaVA, an agricultural multimodal conversation system.
To accelerate progress in this field and inspire more researchers to engage, we design a diverse and challenging evaluation benchmark for agricultural pests and diseases.
Experimental results demonstrate that Agri-LLaVA excels in agricultural multimodal conversation and visual understanding, providing new insights and approaches to address agricultural pests and diseases. 
By open-sourcing our dataset and model, we aim to promote research and development in LMMs within the agricultural domain and make significant contributions to tackle the challenges of agricultural pests and diseases.
All resources can be found at \url{https://github.com/Kki2Eve/Agri-LLaVA}.
\end{abstract}

\begin{figure*}[t]
    \centering
    \includegraphics[width=0.42\textwidth]{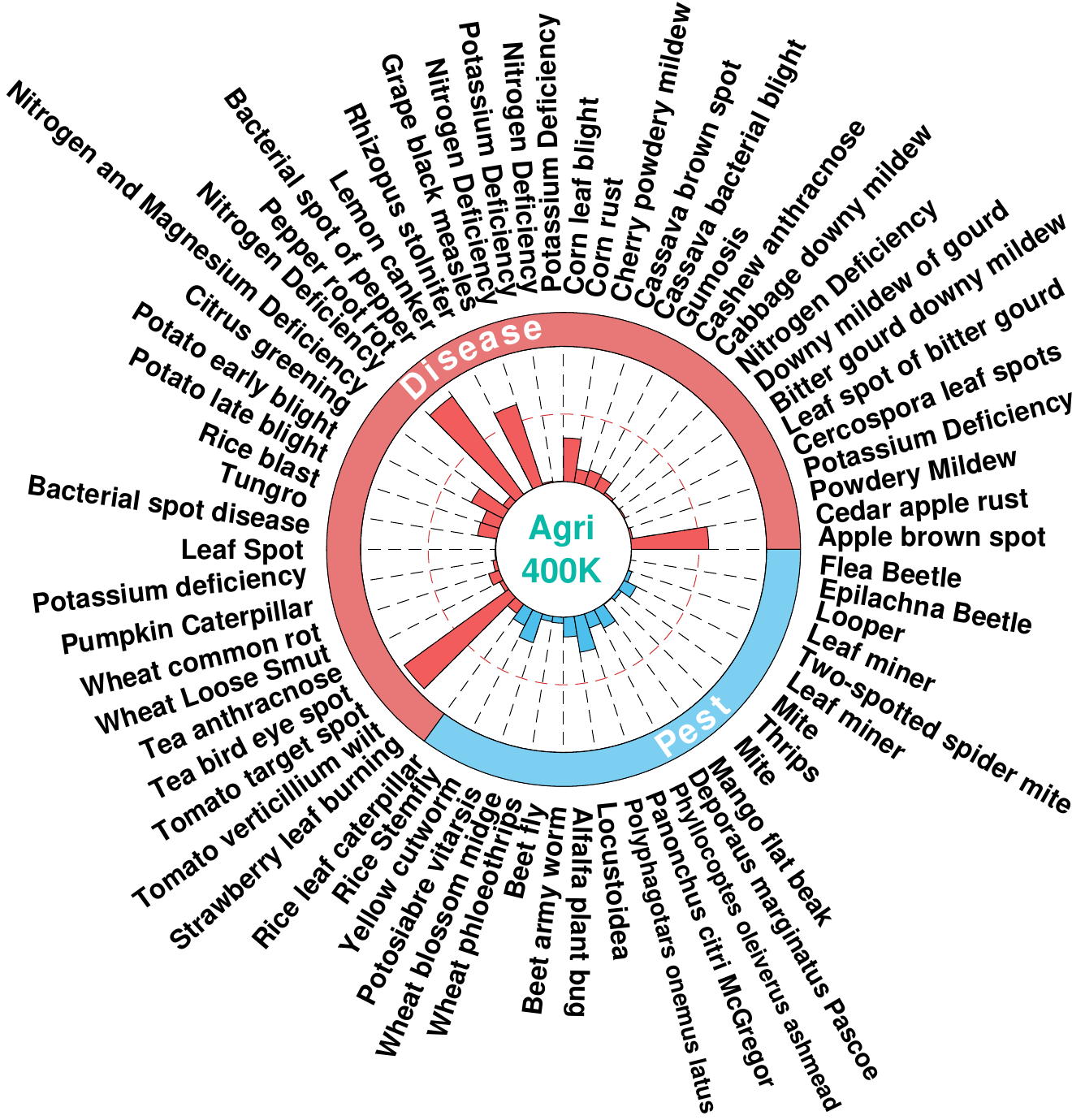}
    \hspace{3em}
    \includegraphics[width=0.42\textwidth]{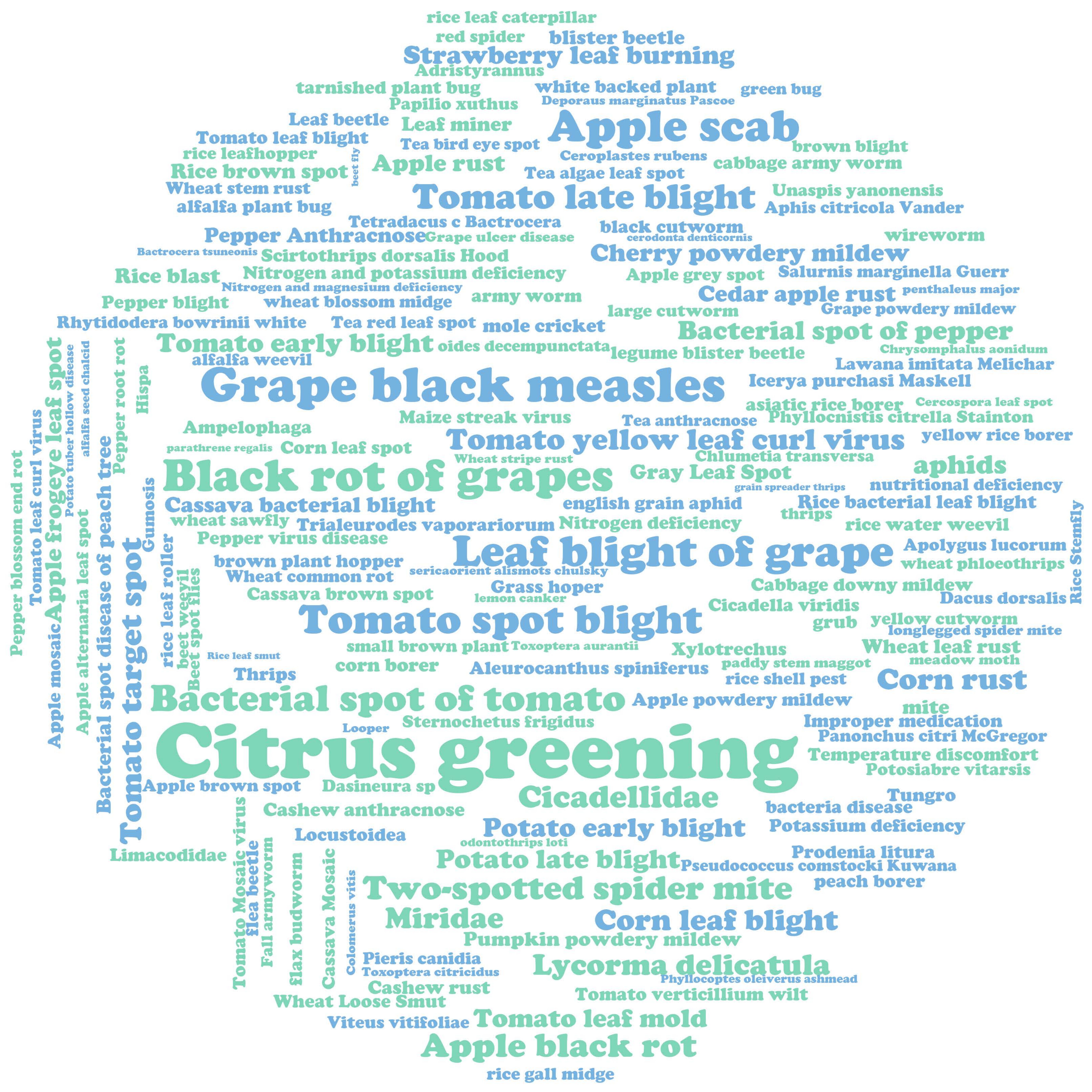}
     \\
    \makebox[0.42\textwidth]{\small (a) Taxonomic system}
    \hspace{3em}
    \makebox[0.42\textwidth]{\small (b) Word cloud distribution}
    \caption{The data statistics of our agricultural multimodal instruction-following data.}
    \label{data distribution}
\end{figure*}

\vspace{-0.5em}
\section{Introduction}
\label{sec:intro}

In recent years, large multimodal models (LMMs)~\cite{alayrac2022flamingo,liu2024visual,dai2024instructblip,zhu2023minigpt,bai2308qwen,lai2023lisa,wang2023cogvlm,you2023ferret} have garnered widespread attention from the research community. 
Compared to unimodal large language models (LLMs)~\cite{mann2020language,chowdhery2023palm,le2023bloom,touvron2023llama,taori2023stanford}, LMMs better mimic human cognitive processes, which understand the world through the collaboration and integration of various sensory inputs. 
Multimodal perception is a crucial component on the path toward achieving artificial general intelligence (AGI)~\cite{bubeck2023sparks}. 
We aim for models to emulate human-like contextual comprehension and adeptly address various tasks with minimal or even no guidance.
Recent research has focused on visual instruction tuning. 
With carefully crafted multimodal instruction-following data, LMMs have demonstrated remarkable task-completion abilities across general domains.

Although LMMs have made significant advancements in the general domain, their application in specific areas, particularly agriculture, faces numerous challenges.
Agricultural pests and diseases, critical issues in agricultural production, present distinct challenges. 
In contrast to general images, agricultural images are inherently more complex, incorporating a greater variety of environmental variables and biological features.
Moreover, identifying and controlling agricultural pests and diseases demand extensive domain-specific knowledge. 
Factors such as rapid spread, strong resistance, and environmental complexity further exacerbate the difficulty of control. 
While the success of LMMs in the medical domain~\cite{li2024llava} has shown the feasibility of fine-tuning for specific domains, agriculture encounters challenges such as data scarcity, unstable data quality, and the need for specialized knowledge. 
These challenges severely impede the development of LMMs for addressing agricultural pests and diseases.
Therefore, addressing these challenges and promoting research and development of LMMs for agricultural pests and diseases is an urgent imperative.

In this paper, we introduce the first multimodal instruc-tion-following dataset specifically designed for the agricultural domain, focusing on identifying agricultural pests and diseases.
By leveraging publicly available datasets~\cite{wu2019ip102,hughes2015open,goyal2021leaf}, competition platforms like Kaggle\footnote{https://www.kaggle.com/} and Baidu PaddlePaddle\footnote{https://aistudio.baidu.com/datasetoverview}, and the Chinese Academy of Agricultural Sciences Pests and Diseases Database\footnote{https://www.cgris.net/disease/}, we assembled a substantial collection of image-text pairs and expert agricultural knowledge pertaining to pests and diseases.
Building upon this foundation, we embarked on large-scale research into multimodal pre-training specifically for recognizing agricultural pests and diseases. 
Guided by extensive agricultural knowledge, our efforts yielded: (i) a dataset of 391,785 image-text pairs related to agricultural pests and diseases, and (ii) a curated dataset of multimodal knowledge-based instruction-tuning conversations specific to agriculture.
This comprehensive dataset lays the groundwork for the development of agricultural multimodal assistants.

Based on our instruction-following dataset, we introduce a novel end-to-end fine-tuning approach for LMMs, marking the first attempt to extend LMMs into the agricultural domain.
Inspired by visual instruction tuning, our knowledge-tuning process is divided into two phases. 
The first phase fine-tunes the model on a vast array of image-text pairs depicting agricultural pests and diseases, aligning agricultural images with corresponding names to enable accurate identification. 
The second phase fine-tunes the model on a dataset of agricultural multimodal knowledge-based conversations, training it to comprehend agricultural queries and respond accurately, thus imbuing it with multimodal conversation capability specific to agriculture. 
Given the specialized nature of agriculture, we incorporate professional agricultural knowledge into both phases to ensure accuracy and mitigate misconceptions. 
Through this process, we unveil Agri-LLaVA, the first agricultural multimodal assistant.

To the best of our knowledge, there is currently no publicly available visual question answering (VQA) dataset specifically for the agricultural domain. 
To foster development and stimulate research interest in this field, we introduce a diverse and challenging benchmark for agricultural pests and diseases. 
This benchmark includes both multimodal chatbot and VQA components. 
We hope our efforts will pave the way for future research endeavors in this domain. 
In summary, our paper makes the following contributions:

\begin{itemize}
    \item \textit{Agricultural multimodal instruction-following data}. 
    We construct the first agricultural multimodal instruction-following dataset, encompassing a wide range of pests and diseases and embedding extensive professional agricultural knowledge.
    \item \textit{Agricultural multimodal assistant}. 
    We propose the first agricultural multimodal assistant, extending LMMs into the agricultural domain through end-to-end fine-tuning enriched with knowledge injection.
    Experimental results confirm the effectiveness of our knowledge-tuning approach, as Agri-LLaVA demonstrates outstanding abilities in completing agricultural multimodal tasks.
    \item \textit{Agricultural multimodal instruction-following benchmark}.
    We introduce the first agricultural multimodal instruction-following benchmark, designed to evaluate the capabilities of agricultural LMMs in tasks such as conversation completion, comprehension, and inference.
    \item \textit{Open-source}.
    To foster community development, all resources will be open-sourced.
\end{itemize}

\begin{figure*}[t]
    \centering
    \includegraphics[width=0.85\linewidth]{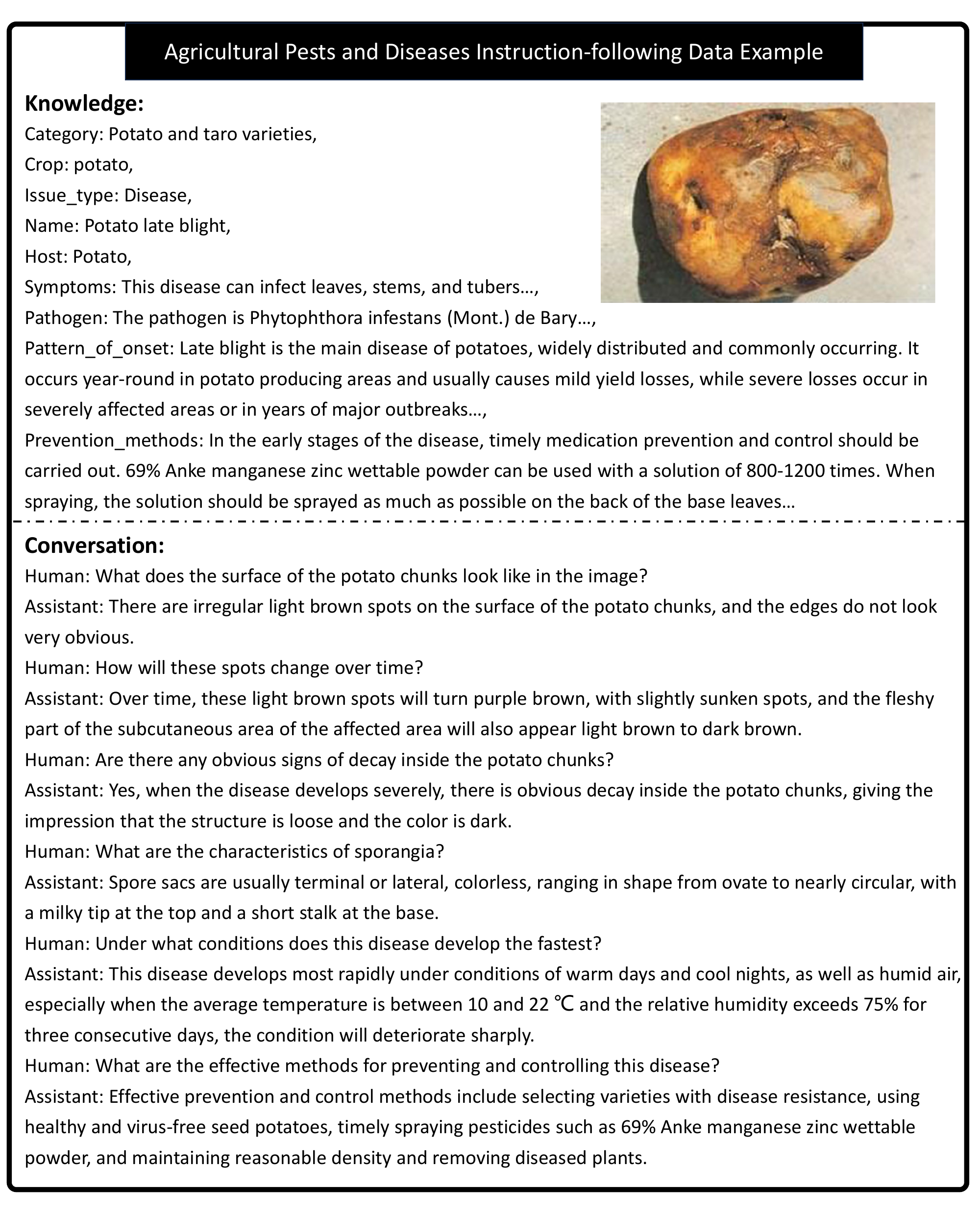}
    \caption{An example of our agricultural pests and diseases instruction-following data. At the top are the image along with its corresponding structured knowledge. At the bottom is the instruction-following data generated by GPT-4 based solely on the provided knowledge.}
    \label{instruction_following_data_example}
\end{figure*}

\section{Related Work}
\label{sec:related works}

\noindent \textbf{Multimodal instruction-following data.}
High-quality instruction-following data has a significant impact on the performance of instruction-following models~\cite{zhou2024lima}.
Existing methods for constructing multimodal instruction-following data roughly fall into three categories. 
The first method is data
adaptation~\cite{dai2024instructblip,wang2024visionllm,chen2023x,xu2022multiinstruct,zhang2023llama,zhao2023chatbridge,luo2024cheap,gao2023llama}, which involves naturally transforming existing image-text pairs datasets (such as VQA datasets) into multimodal instruction data. 
However, data obtained through this method lacks multi-turn conversations, failing to meet real-world application needs. 
To address this gap, LLaVA~\cite{liu2024visual} proposes a method that solely utilizes language models to create multimodal instruction-following data. 
The second method, known as self-instruct~\cite{liu2024visual, maaz2023video,li2024llava,lyu2023macaw,pi2023detgpt,zhu2023minigpt,zhao2023chatbridge}, operates on this idea by utilizing images with detailed captions and bounding boxes. This enables the language teacher model to generate new multimodal data based on contextual information.
Recently, with the release of GPT-4V~\cite{openai2023gpt-4v}, some researchers have opted to use its powerful multimodal capabilities to generate higher quality multimodal data~\cite{wang2023see,chen2024allava}.
The third method is hybrid composition~\cite{ye2023mplug,gong2023multimodal,luo2024cheap,gao2023llama}, which attempts to compensate for the lack of multimodal conversation data by leveraging only language-based conversation data. 
By randomly sampling both language-only data and multimodal data and combining them according to specific methods, this approach integrates single-modal and multimodal data for training, enhancing the instruction-following and conversation capabilities of LMMs.

\noindent \textbf{Instruction-following LMMs.} Utilizing high-quality instruction-following data, instruction-tuning aligns LLMs with human intent, effectively enhancing their few-shot and zero-shot generalization capabilities~\cite{wei2021finetuned}. 
This technique has seen tremendous success in natural language processing (NLP) and has been applied to state-of-the-art LLMs. 
LLaVA~\cite{liu2024visual} extends this technique to the vision-language (VL) multimodal space, developing a general-purpose VL assistant in the process. 
Recent research on LMMs has focused on bridging visual encoders and LLMs using learnable interfaces and training VL assistants through visual instruction tuning. 
Depending on the type of learnable interface used, common LMMs can be classified into three categories:
The first category is query-based interfaces, represented by methods such as Video-LLaMA~\cite{zhang2023video}, MultiModal-GPT~\cite{gong2023multimodal}, mPLUG-Owl~\cite{ye2023mplug}, and MiniGPT-4~\cite{zhu2023minigpt}. 
These methods utilize a set of learnable query tokens to bridge vision and language, learning multimodal information based on queries.
The second category is projection-based methods, which align image features with the semantic space of LLMs using linear layers or multilayer perceptrons, thus narrowing the gap between modalities. 
Representative methods include LLaVA~\cite{liu2024visual}, LLaVA-Med~\cite{li2024llava}, LAMM~\cite{yin2024lamm}, Video-ChatGPT~\cite{maaz2023video}, and PandaGPT~\cite{su2023pandagpt}.
The final category involves methods based on parameter-efficient tuning, where adapter modules are inserted to dynamically learn and allocate weights and information for VL multimodalities, facilitating deep interaction and fusion among different modalities.
Noteworthy methods in this category include LaVIN~\cite{luo2024cheap}, LLaMA-Adapter~\cite{zhang2023llama}, and LLaMA-Adapter V2~\cite{gao2023llama}.

\begin{figure*}[t]
    \centering
    \includegraphics[width=0.9\linewidth]{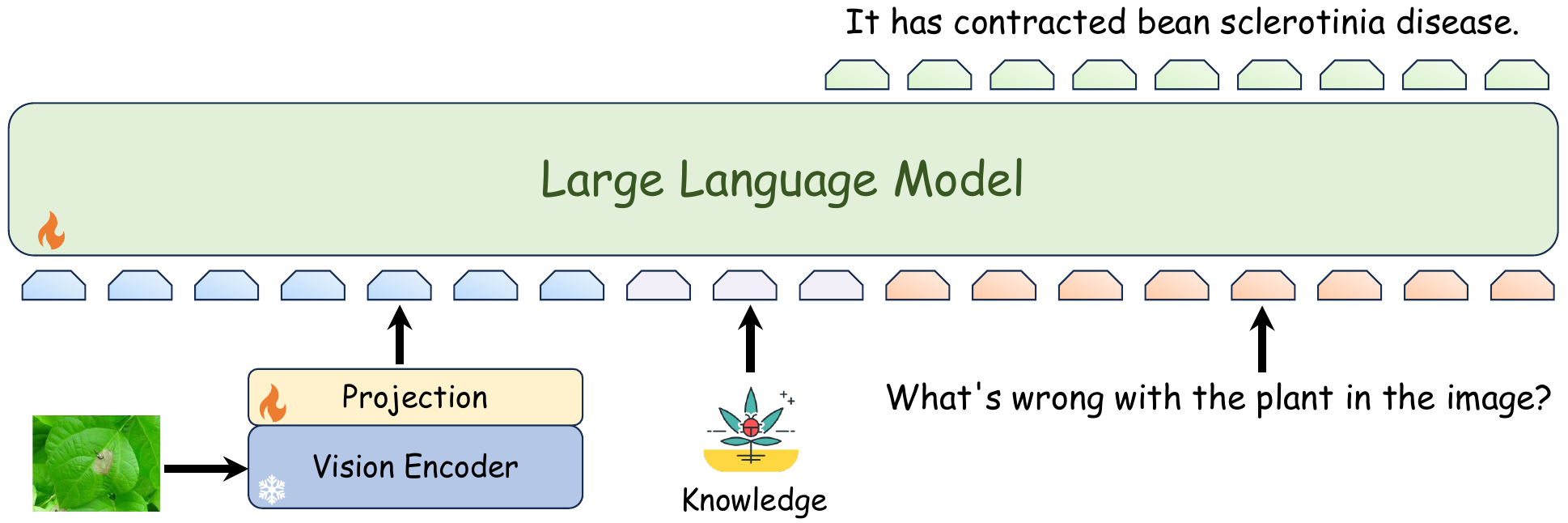}
    \caption{Agri-LLaVA network architecture.}
    \label{Network architecture}
\end{figure*}

\section{Agricultural Instruction-Following Data}
\label{sec:data}

\subsection{Agricultural feature alignment data}
To adapt LMMs from the general domain to the agricultural domain, we use GPT for data construction. 
Utilizing existing publicly available datasets on pests and diseases, we create an agricultural pests and diseases feature alignment dataset consisting of approximately 400,000 samples.
Specifically, we download and preprocess 391,785 images from 16 datasets (see Appendix), including IP102~\cite{wu2019ip102}, which contains 109 disease categories and 112 pest categories. 
Using the pest and disease category labels from the dataset, we search online for corresponding knowledge, retaining category names and detailed symptom descriptions as associated knowledge. The data statistics is shown Figure~\ref{data distribution}.

To enable the model to correlate image features with specific categories, we design distinct instruction templates for various pests and diseases. 
Our dual objectives are: first, to enable the model to recognize image features and thereby identify pest and disease categories. 
Second, to input symptom knowledge, allowing the model to learn detailed symptoms of each pest and disease.
This approach establishes connections between images, categories, and symptoms. 

For each image of pests and diseases $X_i$ and its corresponding knowledge $X_k$, we randomly sample two questions $X_{q1}$ and $X_{q2}$ from the templates.
$X_{q1}$ asks the model to provide a simple description of the pest or disease features in the image, while $X_{q2}$ inquires about the corresponding category of the pest or disease in the image and their detailed symptoms. 
Based on the triplet of (image, question, knowledge), we construct examples of two-round feature alignment conversation:
\vspace{-1em}
\begin{align*}
    Human:X_{q1},X_i\textcolor{Green}{<STOP>}Assistant:X_k\textcolor{Green}{<STOP>} 
    \\
    Human:X_{q2},X_i\textcolor{Green}{<STOP>}Assistant:X_k\textcolor{Green}{<STOP>}
\end{align*}

\subsection{Agricultural instruction-tuning data}
To become a competent multimodal assistant, merely identifying agricultural pests and diseases is insufficient.
It must also possess domain-specific conversational abilities.
To achieve this, we collect 5,813 images of crops infected with pests and diseases, along with corresponding agricultural knowledge sourced mainly from websites such as the Chinese Academy of Agricultural Sciences Pests and Diseases Database. 
We then use GPT-4 to generate professional knowledge-based conversations about these images.

Specifically, we extract agricultural knowledge from web pages and segment the text based on keywords such as symptoms, pathogens, transmission conditions, and control methods. 
This structured knowledge is then organized into a standardized format and stored in a JSON file. 
In this manner, we obtain pairs of images and corresponding agricultural knowledge texts. 
Given the highly knowledge-centric nature of the agricultural domain, we use the structured agricultural knowledge base to guide GPT-4 in generating multi-turn knowledge conversations about the images. 
This approach helps reduce knowledge-related errors in the generated conversation data. 
Additionally, we manually create samples of instruction data to assist GPT-4 in understanding how to generate compliant instruction-following data. 
Through these processes, we ultimately obtain 6,000 high-quality agricultural multimodal conversation data. 
An example is shown in Figure~\ref{instruction_following_data_example}.

\section{Modeling}
\label{sec:finetune}

We use LLaVA-1.5~\cite{liu2024improved} as the base model to train our agricultural multimodal assistant. Our entire training process is divided into two stages, and the overall model architecture is illustrated in Figure~\ref{Network architecture}.

\subsection{Pre-training for Feature Alignment}
During this stage, we primarily utilize the agricultural pests and diseases feature alignment data introduced in Section~\ref{sec:data}. 
Throughout the training process, we keep the weights of the visual encoder and LLM frozen, training only the projection matrix within the model. 
Given images of agricultural pests and diseases, we task the model with accurately predicting the specific type of disease or pest and providing detailed symptom descriptions of the identified disease or pest. 
The objective of this stage is to enable the model to establish correspondence between the features of agricultural pests and diseases images, detailed symptom descriptions, and their respective categories, thereby endowing the model with the ability to identify agricultural pests and diseases.

\subsection{End-to-End Instruction-tuning}
In this stage, we utilize the agricultural pests and diseases instruction-tuning data introduced in Section~\ref{sec:data}. 
During training, we freeze only the weights of the visual encoder while updating both the projection matrix and the LLM's weights. 
Following pre-training, the model acquires a certain level of domain-specific knowledge in agriculture but lacks the ability to answer questions. 
By fine-tuning the model on diverse conversational data, we align the model with human intent, enabling it to address and respond to relevant domain-specific questions. 
This process results in the development of an interactive agricultural multimodal assistant capable of engaging with users.

\begin{table}
    \centering
    \begin{tabular}{cc|c|c|c}
        \toprule[1.5pt]
        \multicolumn{2}{c|}{\multirow{2}{*}{\textbf{Model}}} & \multicolumn{2}{c|}{\textbf{Issue Types}} & \multirow{2}{*}{\textbf{Overall}}\\
        & & Disease & Pest & \\
        \midrule
        \multicolumn{2}{l|}{LLaVA} & 52.2 & 43.6 & 50.6 \\
        \midrule
        \multicolumn{2}{l}{\textbf{\textit{Variants of Agri-LLaVA}}}\\
         \multicolumn{2}{l|}{Pretrain} & 30.9 & 33.2 & 31.3\\
         \multicolumn{2}{l|}{1K lora-finetune} & 46.6 & 47.5 & 46.7 \\
         \multicolumn{2}{l|}{1K full-finetune} & 47.5 & 46.4 & 47.3 \\
         \multicolumn{2}{l|}{6K lora-finetune} & 52.4 & 54.5 & 52.8 \\
         \multicolumn{2}{l|}{6K full-finetune} & 53.2 & 59.6 & 54.4 \\
         \rowcolor{gray!30}
         \multicolumn{2}{l|}{6K + extra 3K finetune} & \textbf{54.0} & \textbf{61.4} & \textbf{55.4} \\
        \bottomrule[1.5pt]
    \end{tabular}
    \caption{The results on Agri-LLaVA-Chatbot-Bench, evaluated by the relative scores provided by GPT-4, quantitatively measure the model's instruction-following abilities. We report the outcomes of Agri-LLaVA ablations, conducted with varying training data and methodologies.}
    \label{chatbot quantitative analysis}
\end{table}

\begin{table*}[t]
    \centering
    \renewcommand{\arraystretch}{1.1}
    \small
        \begin{tabularx}{0.9\textwidth}{lX}
        \toprule[1.5pt]
        \textbf{Visual input example} & \\
        \midrule
        & $\raisebox{.03\height}{\includegraphics[width=0.5\linewidth]{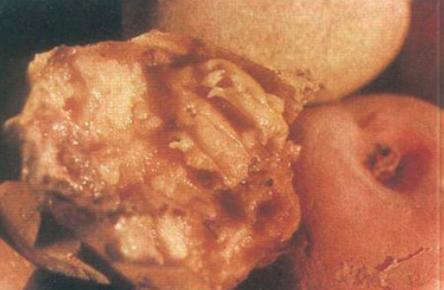}}$  \\
        User & What is the crop in this image? \\
        Agri-LLaVA & The crop in the image is potatoes. \\
        User & How to prevent and treat this condition? \\
        Agri-LLaVA & The prevention and control methods include using disease-free seeds, implementing crop rotation, and strengthening field management. In the early stages of the disease, specific pesticides can be used for spraying, such as 75\% chlorothalonil wettable powder 600 times liquid, 50\% benzimidazole emulsion 1500 times liquid, and 50\% carbendazim wettable powder 800 times liquid, etc. \\
        \midrule
        User & What is the crop in this image? \\
        LLaVA & The crop in this image is a pig's ear. \\
        User & How to prevent and treat this condition? \\
        LLaVA & The image shows a close-up of a pig's ear with a hole in it. This condition is likely caused by an infection or injury. To prevent and treat this condition, it is essential to maintain proper hygiene and care for the pig. \\
        \midrule
        User & What is the crop in this image? \\
        Language-only GPT-4 & The crop in this image is potatoes. \\
        User & How to prevent and treat this condition? \\
        Language-only GPT-4 & To prevent and treat potato soft rot disease, the first priority is to strengthen field management, pay attention to ventilation and light transmission, and reduce field humidity. In addition, promptly remove diseased plants and disinfect the affected areas with lime to reduce the source of initial and recurrent infections in the field. In addition, flooding should be avoided. When necessary, pesticides such as 50\% Bacillus thuringiensis wettable powder 500 times solution or 12\% Green Copper emulsion 600 times solution, 47\% Garenon wettable powder 500 times solution, and 14\% Chloramphenicol water solution 300 times solution can be sprayed to control the condition. \\
        \bottomrule[1.5pt]
    \end{tabularx}
    \caption{The visualization results of the multimodal instruction-following capabilities in Agri-LLaVA-Chatbot-Bench. Compared to LLaVA, Agri-LLaVA precisely identifies crop types and diagnoses diseases based on image features. Responses generated by the language-only GPT-4 based on the knowledge base are considered the performance upper limit.}
    \label{chatbot qualitative analysis}
\end{table*}

\section{Experiment}
\label{sec:experiment}

Our experiments primarily evaluate two key capabilities of Agri-LLaVA: instruction following and visual reasoning. 
Consequently, the experiments are divided into two parts: testing the multimodal chatbot and testing VQA. 
We address two main questions: (1) Is our data quality sufficient to support our model as an agricultural multimodal assistant? 
(2) Does our agricultural multimodal assistant achieve the expected performance level?

Our model is trained on 8 A800 GPUs, with the entire training process taking 11 hours and 20 minutes. 
Initially, we pre-train the model using 400K feature alignment data with a learning rate of 1e-3 and a batch size of 256 for 1 epoch, which takes 10 hours and 40 minutes. 
Subsequently, we fine-tune the model using 6K instruction-following data with a learning rate of 2e-5 and a batch size of 128 for 3 epochs, which takes 40 minutes. 
It's worth noting that throughout the entire experimental process, we solely utilize the language-only GPT-4.

\subsection{Agricultural Multimodal Chatbot}

\subsubsection{Agri-LLaVA-Chatbot-Bench}
To evaluate Agri-LLaVA's instruction-following ability, we randomly select 30 images of various pests and diseases from Baidu Baike\footnote{https://baike.baidu.com/} and World Agrochemicals Network\footnote{https://cn.agropages.com/}. 
This set includes 6 images of pests and 24 of diseases. 
To test Agri-LLaVA's performance on more challenging tasks and its generalization ability in unseen scenarios, we deliberately choose 25 types of pests and diseases not encountered during the training process.
Using the same instruction-following data generation pipeline as in the second stage, we generate 4 to 6 rounds of conversation per image. 
These conversations cover various aspects of pest and disease knowledge, including symptoms, pathogens, transmission, and control, aiming to comprehensively evaluate Agri-LLaVA's understanding and execution capabilities.
Ultimately, we generate 151 rounds of conversation, providing ample data to support the evaluation of the model's performance.
This experimental design enables a comprehensive and accurate assessment of Agri-LLaVA's ability to understand and follow instructions, as well as its generalization capability in handling unseen pest and disease scenarios.

\subsubsection{Evaluation Criteria}
To assess and understand the multimodal conversation capability of Agri-LLaVA, we use GPT-4 to quantify the model's accuracy in answering questions. 
Specifically, we create triplets of (image, question, knowledge), where GPT-4 answers questions based on the provided knowledge.
We use its responses as the theoretical performance limit, serving as the ground truth for the question. 
Then, we task candidate models with answering the same questions based on the images. 
After obtaining responses from both the candidate model and GPT-4 for the same image and question, we input the image, question, knowledge, and responses from both assistants into GPT-4. 
We then ask it to evaluate the helpfulness, relevance, accuracy, and level of detail of the responses from the two assistants, assigning a relative score ranging from 1 to 10. 
A higher score indicates a better response, implying superior model performance. 
Additionally, we request detailed explanations from GPT-4 regarding the evaluation, aiding in better comprehension of the model's performance in this task. 
This evaluation method comprehensively and objectively assesses the model's abilities in multimodal conversation tasks, providing crucial insights for model refinement.

\subsubsection{Quantitative Analysis}
Quantitative analysis results, as shown in Table~\ref{chatbot quantitative analysis}, reveal that Agri-LLaVA pretrained solely on stage-1 data exhibits subpar performance in instruction-following and lacks diversified conversational abilities. 
However, significant improvements in conversational capabilities are observed when a portion of stage-2 data is used to fine-tune Agri-LLaVA. 
As the volume of training data increases, Agri-LLaVA's performance gradually improves. 
When instruction-following data reaches 6,000, Agri-LLaVA's performance surpasses that of LLaVA. 
Furthermore, the addition of 3,000 simple single-round conversation data to Agri-LLaVA further enhances performance, indicating the critical importance of high-quality agricultural instruction-following data in developing agricultural multimodal assistants. 
Experimental results demonstrate that our Agri-LLaVA achieves 55.4\% of GPT-4's performance. 
Nonetheless, we believe that with the infusion of more agricultural knowledge, our model will exhibit even better and more professional performance.

\subsubsection{Qualitative Analysis}
Table~\ref{chatbot qualitative analysis} presents the results of the qualitative analysis. 
With the infusion of extensive agricultural knowledge, Agri-LLaVA has demonstrated a certain level of ability to comprehend images and engage in reasoning. 
Compared to LLaVA, Agri-LLaVA can integrate image features with learned agricultural expertise to identify disease types and provide corresponding prevention and control suggestions. 
Although Agri-LLaVA's responses lack the level of detail seen in GPT-4, such as specific recommendations regarding ventilation, light transmission, field humidity, removal of diseased plants, and disinfection of affected areas, they remain accurate and useful. 
This indicates that our training approach is a viable method for developing agricultural multimodal assistants and provides a reliable foundation for Agri-LLaVA's practical applications.

\subsection{Agricultural VQA}

\subsubsection{Agri-LLaVA-VQA-Bench}
As far as we know, there is currently no publicly available dataset specifically for agricultural pests and diseases VQA. 
To test the model's visual reasoning abilities regarding pests and diseases, we randomly select 49 types of diseases, 50 types of pests, and some healthy samples from existing publicly available datasets, totaling 482 images. 
Among these images, there are 6 healthy samples and 476 samples of pests and diseases. 
When selecting the images, we follow certain principles: first, we prioritize choosing types of pests and diseases that do not appear during the training process. 
Second, we ensure the selection of images that do not appear in the training data, ensuring the fairness and effectiveness of the test. 
Ultimately, in our dataset, there are 21 types of diseases and 3 types of pests that do not appear during training.
After careful selection, we manually annotate each image to generate corresponding question-answer pairs.
To thoroughly assess the model's visual reasoning abilities, we design 4-5 rounds of conversation for each image, resulting in a total of 2,268 question-answer pairs. 
These questions cover various aspects of pest and disease damage to organs, abnormal symptoms, related attributes, potential hazards, nomenclature, causes of occurrence, prevention and control methods, transmission routes, and other relevant topics, totaling 9 themes. 
Through these questions, we comprehensively test the model's understanding and reasoning abilities regarding pest and disease images.

\begin{table} 
    \centering
    \resizebox{0.48\textwidth}{!}{
    \begin{tabular}{lccc|cc|c}
        \toprule[1.5pt]
        \multicolumn{4}{c|}{\textbf{Model}} & \multicolumn{2}{c|}{\textbf{Question Types}} & \multirow{2}{*}{\textbf{Average}}\\
        Ins. & St.1 & St.2 & SFT & Open & Closed & \\
        \midrule
        \multicolumn{4}{l}{\textbf{\textit{Variants of Agri-LLaVA}}}\\
        0 & 1 & 0 & 0 & 3.81 & 30.10 & 16.96 \\
        1K & 1 & 3 & 0 & 4.70 & 61.17 & 32.94 \\
        1K & 1 & 3 & 1 & 23.47 & 85.92 & 54.70 \\
        1K & 1 & 3 & 3 & 27.34 & 83.98 & 55.66  \\
        6K & 1 & 3 & 0 & 5.49 & 67.96 & 36.73 \\
        6K & 1 & 3 & 1& 26.01 & 86.89 & 56.45 \\
        \rowcolor{gray!30}
        6K & 1 & 3 & 3& \textbf{30.77} & \textbf{89.32} & \textbf{60.05} \\
        \midrule
        LLaVA~\cite{liu2024improved} & - & - & 3 & 28.32 & 82.04 & 55.18 \\
        Mini-Gemini~\cite{li2024mini} & - & - & 3 & 27.37 & 81.16 & 54.27 \\
        Qwen-VL-Chat~\cite{bai2308qwen} & - & - & 3 & 30.19 & 84.47 & 57.33 \\
        ShareGPT4V~\cite{chen2023sharegpt4v} & - & - & 3 & 29.39 & 85.36 & 57.38 \\
        \bottomrule[1.5pt]
    \end{tabular}}
    \caption{The results on Agri-LLaVA-VQA-Bench. ``Ins.'' is the quantity of instruction-following data. ``St.1'' is the stage 1: pre-training for feature alignment. ``St.2'' is the stage 2: end-to-end instruction-tuning. ``SFT'' is supervised fine-tuning. We report the outcomes of Agri-LLaVA ablations under different conditions.}
    \label{vqa quantitative analysis}
\end{table}

\subsubsection{Evaluation Criteria}
Our VQA evaluation metrics consist of two main components: 
for closed-set questions, we use accuracy to measure the model's ability to provide correct answers within the known question scope.  
For open-set questions, we employ the F1-score~\cite{goutte2005probabilistic} to gauge the accuracy of the responses. 
Open-set questions involve answering queries in unknown domains, so the F1-score better reflects the model's coverage and accuracy across diverse queries.
Together, these two metrics combined comprehensively assess the model's visual reasoning capabilities.

\subsubsection{Ablations}
Table~\ref{vqa quantitative analysis} presents the results of Agri-LLaVA-VQA-Bench, comparing the performance of Agri-LLaVA with the general-domain LMMs and investigate the impact of different instruction-following data constructions and hyperparameters during downstream task fine-tuning. 
Our main findings are as follows:
(1) Models pre-trained with only stage-1 data exhibit significantly weaker visual reasoning capabilities compared to models fine-tuned in stage-2. 
This is attributed to the limitation imposed by a single feature alignment dataset on the model's ability to learn diverse instructions.
(2) Following 3 epochs of supervised fine-tuning on the VQA training set, Agri-LLaVA is better than other general-domain LMMs, especially demonstrates a 4.87\% higher comprehensive ability than LLaVA. 
This suggests the effectiveness of our knowledge-infused approach in adapting a general model to the agricultural domain. 
When performing downstream agricultural tasks, our Agri-LLaVA serves as a more suitable base model.
(3) Performance on downstream tasks increases with the augmentation of stage-2 instruction-following data under the same hyperparameters for supervised fine-tuning. This underscores the crucial impact of high-quality instruction-following data on model performance. 
While the performance of some variants of Agri-LLaVA is surpassed by LLaVA, this is due to the higher difficulty level of Agri-LLaVA-VQA-Bench. 
When the data volume is low, the knowledge acquired by Agri-LLaVA may not sufficiently bridge the zero-shot capability gap between it and the general LLaVA.



\vspace{-0.3em}
\section{Conclusion}
\label{sec:conclusion}
We propose Agri-LLaVA, the first large-scale vision-and-language model specifically tailored for the agricultural domain. 
To train this model, we design and construct a massive
agricultural multimodal instruction-following dataset, integrating extensive knowledge of agricultural pests and diseases with high-quality agricultural conversational data. Additionally, to comprehensively evaluate Agri-LLaVA's capabilities in instruction following and visual reasoning, we introduce the first agricultural multimodal benchmark. 
Experiments demonstrate that Agri-LLaVA exhibits the expected proficiency in agricultural conversational and reasoning tasks. 
We believe that Agri-LLaVA marks an important step forward in the development of large multimodal models for agriculture. However, given the complexity of the agricultural domain, which presents challenges comparable to or even greater than those faced by most LMMs, Agri-LLaVA may still generate inaccuracies and harmful outputs. 
Future work will focus on reducing these illusions and injecting more domain-specific knowledge to enhance the model's capabilities and reliability.

\section*{Acknowledgements}
This work was supported by the National Natural Science Foundation of China (Grant NO. 62122035).

{
    \small
    \bibliographystyle{ieeenat_fullname}
    \bibliography{main}
}

\newpage

\appendix
\section*{Appendix}

Due to space limitations, many details have been omitted in the main text, we provide relevant additional information here.

1. Section~\ref{Data Generation}: More details of data generation.

2. Section~\ref{Data Format}: Data format.

3. Section~\ref{Data Preprocessing}: Data preprocessing.

4. Section~\ref{Data Cleaning}: Data cleaning.

5. Section~\ref{Data Supplement}: Data supplement.

6. Section~\ref{More Results}: More results.

7. Section~\ref{Limitations}: Limitations.

8. Section~\ref{Broader Impact}: Broader impact.

9. Section~\ref{Evaluation Metrics}: Evaluation metrics.

\section{More Details of Data Generation}
\label{Data Generation}

\begin{figure*}
    \centering
    \includegraphics[width=0.85\linewidth]{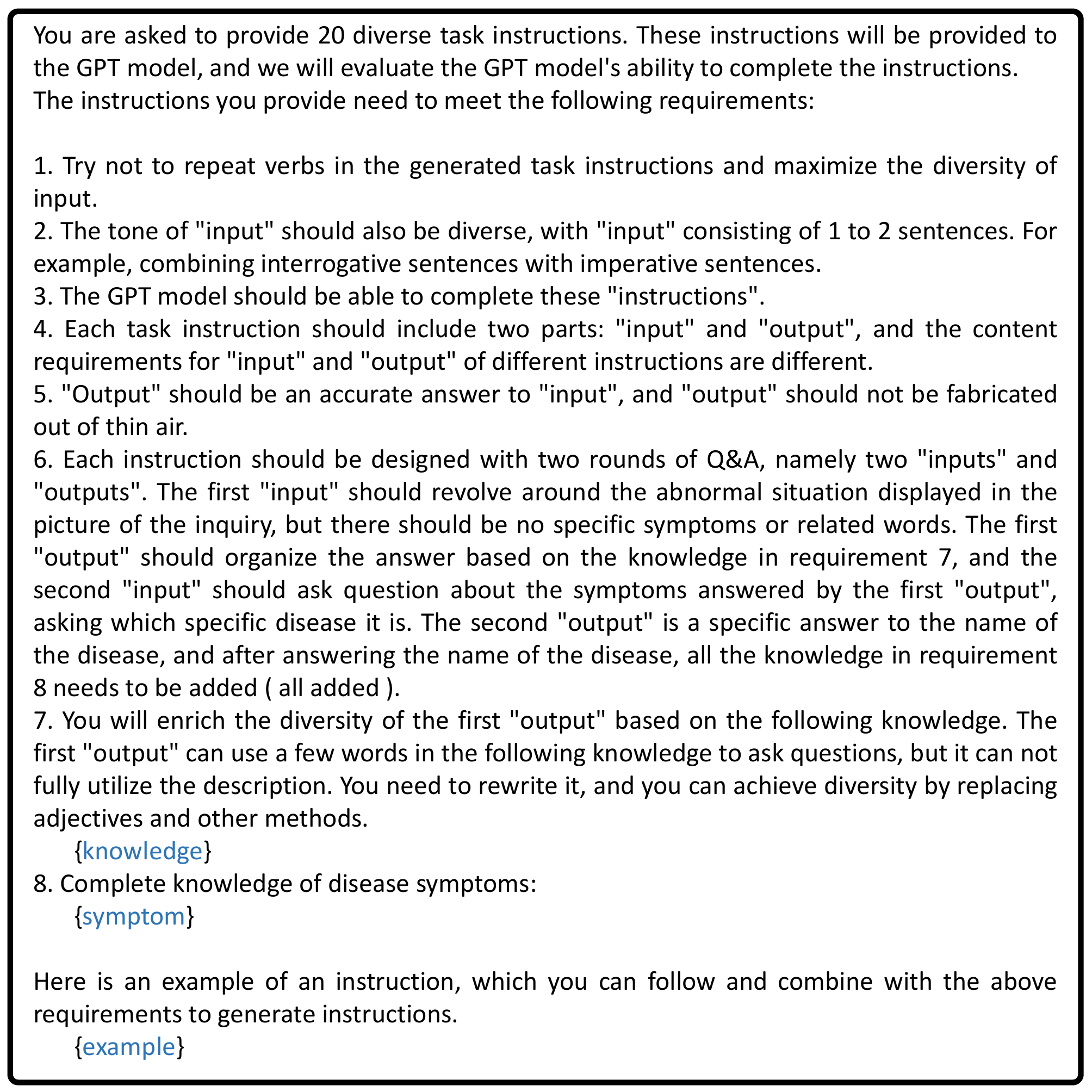}
    \caption{One example of prompt used to generate disease feature alignment data.}
    \label{prompt of stage-1 disease}
\end{figure*}

\textbf{Data source.}
Table~\ref{data source} provides an overview of the public datasets utilized in our study, detailing the sources of data that form the basis for our analysis.

\begin{figure*}[h]
    \centering    
    \includegraphics[width=0.85\linewidth]{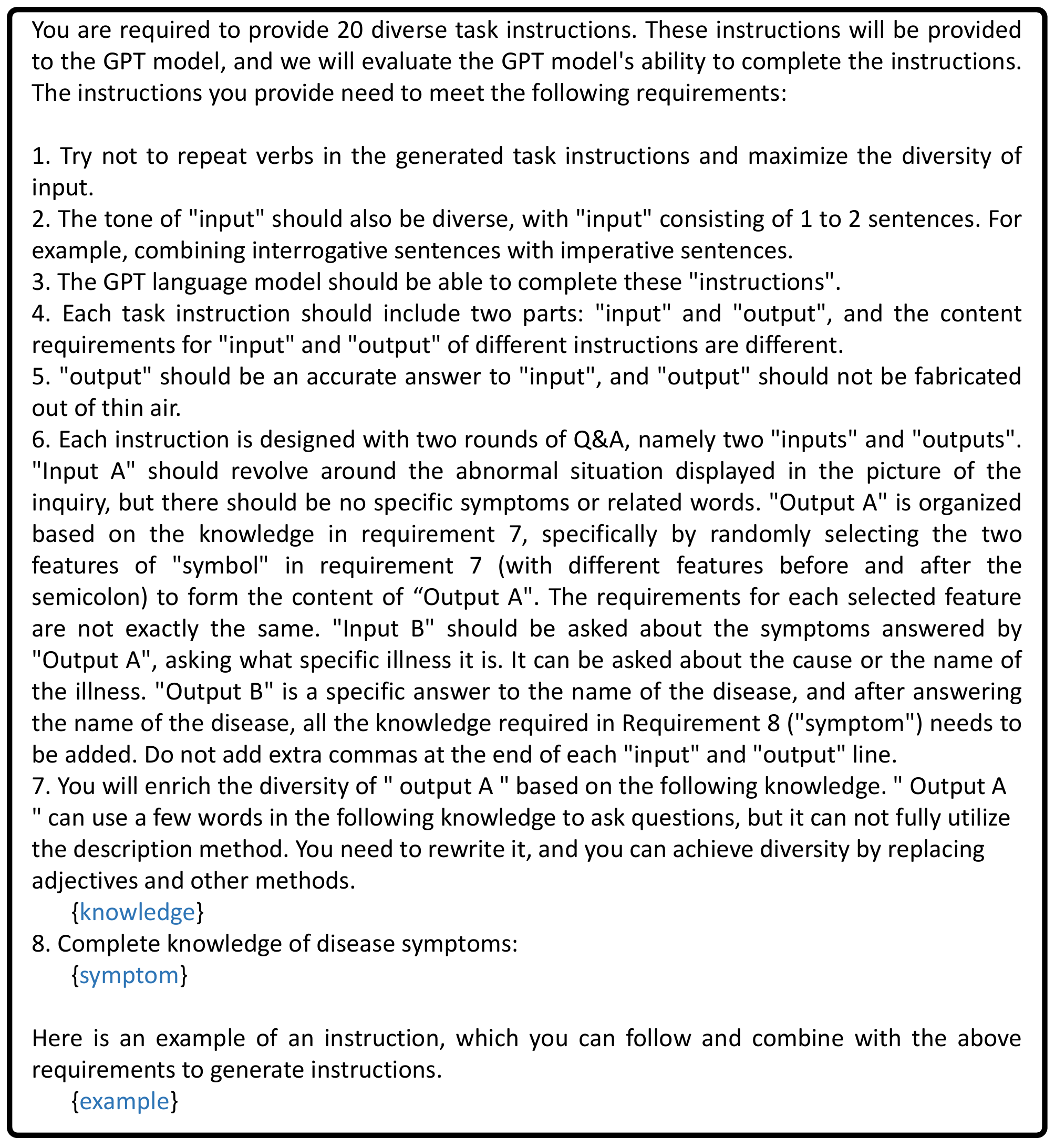}
    \caption{One example of prompt used to generate pest feature alignment data.}
    \label{prompt of stage-1 pest}
\end{figure*}

\textbf{Prompts for feature alignment data.} 
The prompts used to guide GPT-4 to generate feature alignment data from knowledge are shown in Figure~\ref{prompt of stage-1 disease} and Figure~\ref{prompt of stage-1 pest}.

\begin{figure*}[h]
    \centering
    \includegraphics[width=0.85\linewidth]{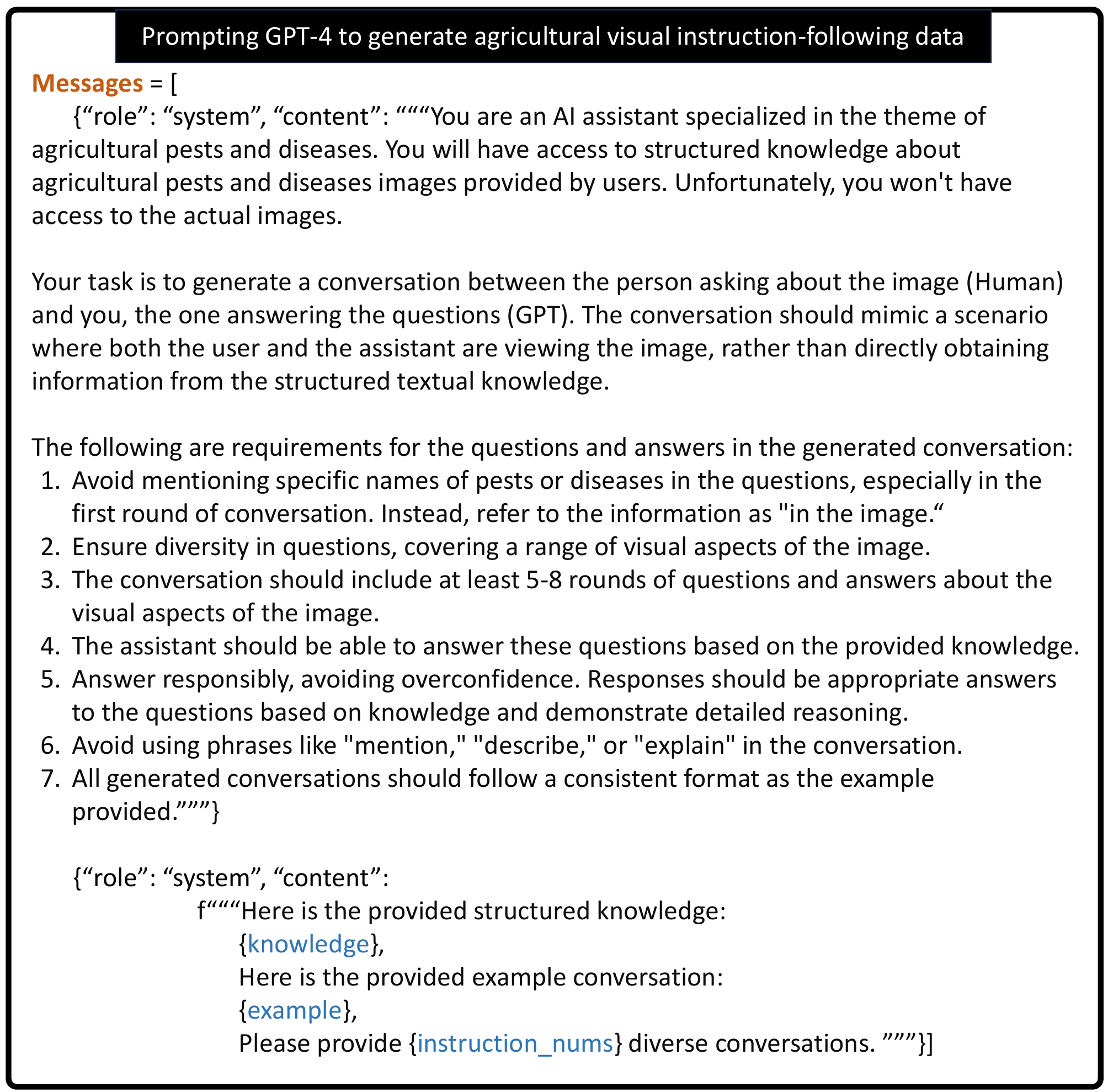}
    \caption{The prompt used to generate instruction-tuning data. The message provides detailed instructions for guiding GPT-4 in generating conversations. Based on the structured knowledge provided, we expect GPT-4 to generate diverse conversations consistent with the format of example conversations.}
    \label{prompt of stage-2}
    \vspace{-0.4em}
\end{figure*}

\textbf{Prompts for instruction-following data.} 
The prompt used to guide GPT-4 to generate instruction-tuning data from knowledge is shown in Figure~\ref{prompt of stage-2}.

\section{Data Format}
\label{Data Format}
The dataset we have constructed is in the form of image-text pairs, where the images are in jpg format and the text is recorded in JSON file. Our dataset can be divided into four parts: agricultural pests and diseases feature alignment data, agricultural pests and diseases instruction-tuning data, Agri-LLaVA-Chatbot-Bench and Agri-LLaVA-VQA-Bench. In these four sub-datasets, the text content is designed around knowledge of agricultural pests and diseases, but there are slight differences in format and content.

\begin{table}
    \centering
    \resizebox{\linewidth}{!}{
    \begin{tabular}{l|c}
    \toprule[1.5pt]
         \textbf{Data} & \textbf{Source} \\
         \midrule
         AppleLeaf9 & \href{https://github.com/JasonYangCode/AppleLeaf9?tab=readme-ov-file}{Link} \\
         LWDCD & \href{https://aistudio.baidu.com/datasetdetail/231774}{Link} \\
         PlantVillage-Dataset-master & \href{https://data.mendeley.com/datasets/tywbtsjrjv/1}{Link} \\
         Rice Leaf Disease Image samples & \href{https://data.mendeley.com/datasets/fwcj7stb8r/1}{Link} \\
         Chinese Academy of Sciences disease sample data & \href{https://aistudio.baidu.com/datasetdetail/57324/0}{Link} \\
         IP102 & \href{https://github.com/xpwu95/IP102/blob/master}{Link} \\
         Images of apple diseases and pests & \href{https://aistudio.baidu.com/datasetdetail/84322}{Link} \\
         Crop Pest and Disease Detection & \href{https://data.mendeley.com/datasets/bwh3zbpkpv/1}{Link} \\
         IDADP-grape disease identification & \href{ https://www.scidb.cn/en/detail?dataSetId=76b39c9c435d4035b5076412c2ddcb61}{Link} \\
         OLID I & \href{https://www.kaggle.com/datasets/raiaone/olid-i}{Link} \\
         Plant Disease Expert & \href{https://www.kaggle.com/datasets/sadmansakibmahi/plant-disease-expert/data}{Link} \\
         RiceLeafs & \href{https://aistudio.baidu.com/datasetdetail/73140}{Link} \\
         Cabbage & \href{https://aistudio.baidu.com/datasetdetail/128016}{Link} \\
         Pepper pests and diseases & \href{https://aistudio.baidu.com/datasetdetail/153190/2}{Link} \\
         Wheat disease classification & \href{https://aistudio.baidu.com/datasetdetail/127559/2}{Link} \\
         Wheat Leaf Dataset & \href{https://data.mendeley.com/datasets/wgd66f8n6h/1}{Link} \\
         \bottomrule[1.5pt]
    \end{tabular}}
    \caption{The data source of our dataset.}
    \label{data source}
\end{table}

\textbf{Agricultural pests and diseases feature alignment data.} Agricultural pests and diseases feature alignment data is designed to help the model associate images with categories of agricultural pests and diseases, as well as to acquire knowledge about these pests and diseases. The JSON data for agricultural pests and diseases feature alignment includes two fields: “image” and “conversations”. “image” represents the image of pest or disease, “conversations” is presented in the form of dialogues to enable the model to grasp knowledge about the pests and diseases corresponding to the given image.

\begin{figure*}[h]
    \centering
    \includegraphics[width=0.85\linewidth]{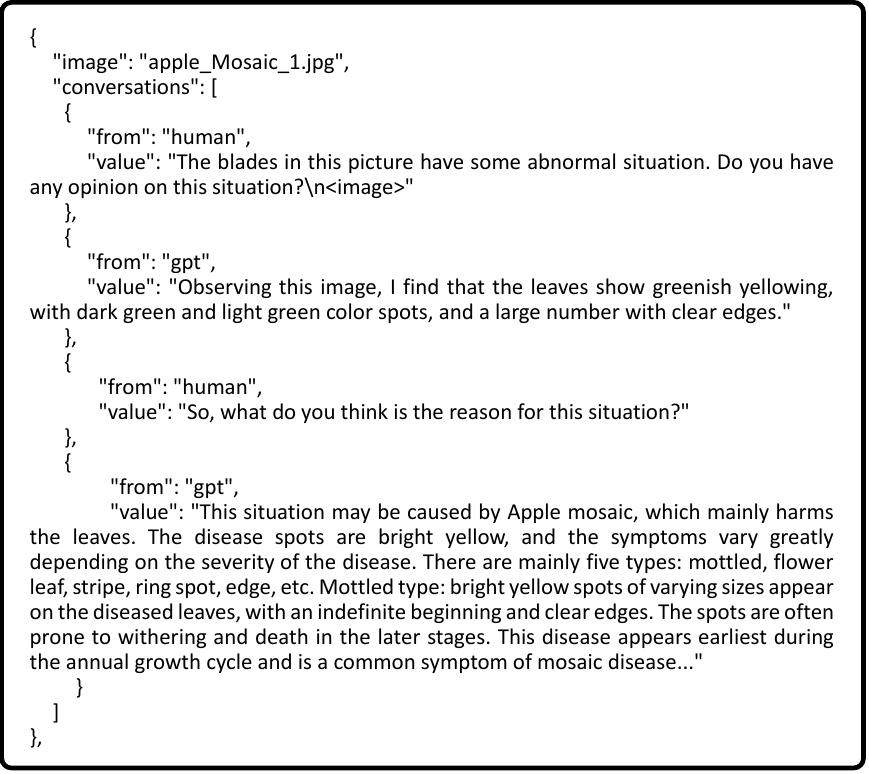}
    \caption{The JSON format of agricultural pests and diseases feature alignment data.}
    \label{The JSON format of agricultural pests and diseases feature alignment data.}
\end{figure*}

\textbf{Agricultural pests and diseases instruction-tuning data.}
Agricultural pests and diseases instruction-tuning data is intended to help the model acquire more agricultural knowledge, such as prevention, transmission methods, etc., rather than just identifying the type of pest or disease. The JSON data for agricultural pests and diseases instruction-tuning includes two fields: ``image'' and ``conversations''. Unlike the ``conversations'' field in agricultural pests and diseases feature alignment data, these ``conversations'' involve more rounds of dialogue and cover a wider range of knowledge about pests and diseases.

\begin{figure*}[h]
    \centering
    \includegraphics[width=0.85\linewidth]{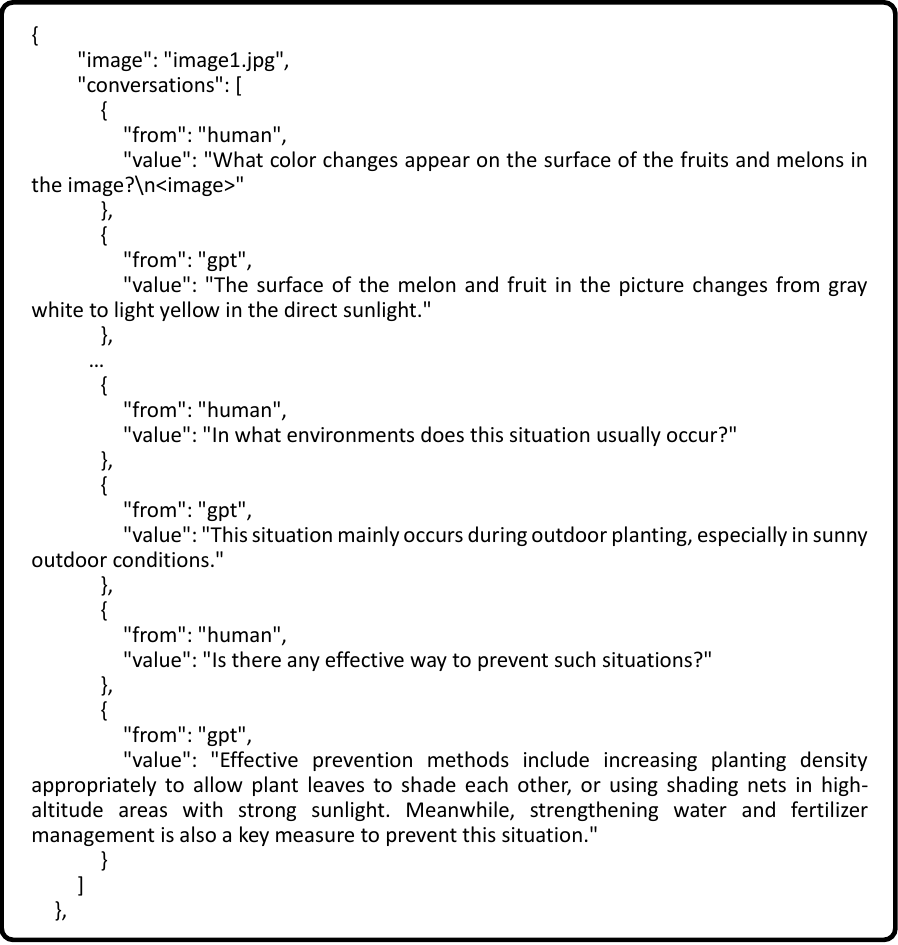}
    \caption{The JSON format of agricultural pests and diseases instruction-tuning data.}
    \label{The JSON format of agricultural pests and diseases instruction-tuning data.}
\end{figure*}

\textbf{Agri-LLaVA-Chatbot-Bench.}
To test the model's ability to execute instructions and generalization capabilities, we designed Agri-LLaVA-Chatbot-Bench. It includes common abnormal phenomena, pathogens, transmission methods, and other issues. Unlike the Agri-LLaVA-VQA-Ben-ch, the answers in the Agri-LLaVA-Chatbot-Bench are significantly longer in length.

\textbf{Agri-LLaVA-VQA-Bench.} 
The Agri-LLaVA-VQA-Bench is designed to test the model's visual reasoning abilities regarding pests and diseases after training. To achieve this, when designing the Agri-LLaVA-VQA-Bench, we considered questions related to identifying pests and diseases, as well as questions about transmission methods, prevention, and other issues. With this purpose in mind, we designed the Agri-LLaVA-VQA-Bench, which includes six fields in its JSON file.

\begin{figure*}[h]
    \centering
    \includegraphics[width=0.85\linewidth]{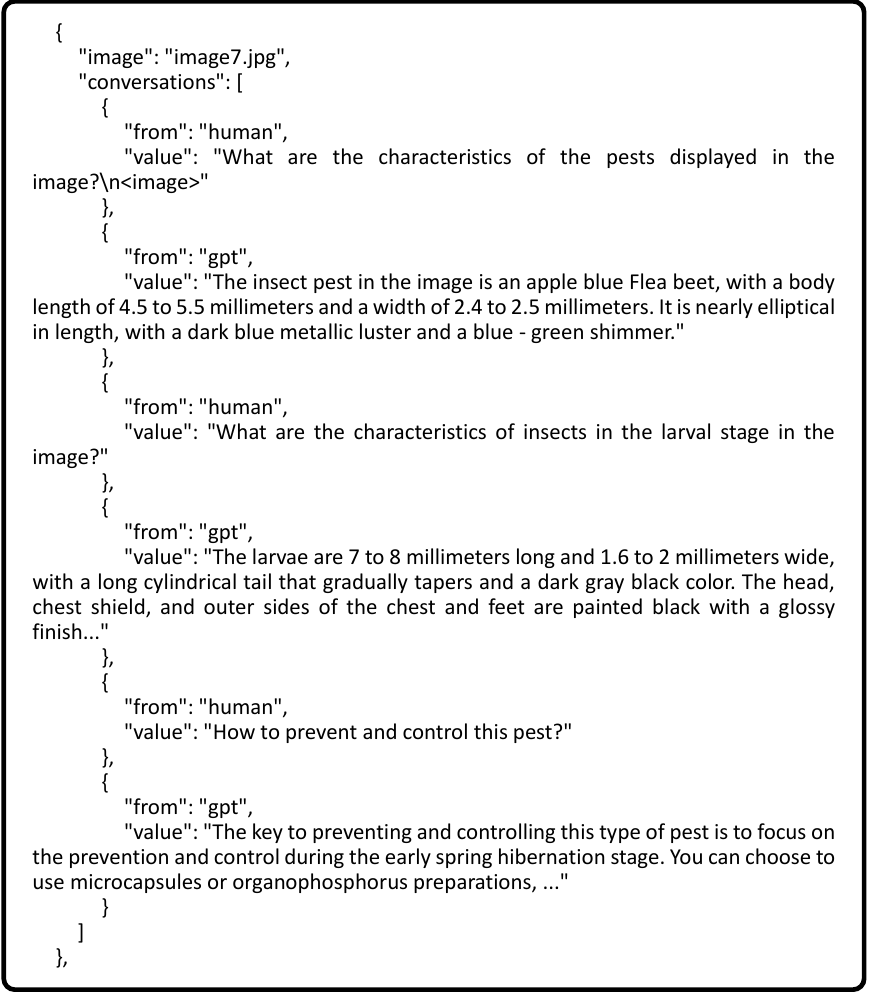}
    \caption{The JSON format of Agri-LLaVA-Chatbot-Bench.}
    \label{The JSON format of Agri-LLaVA-Chatbot-Bench.}
\end{figure*}

\begin{figure*}[h]
    \centering
    \includegraphics[width=0.63\linewidth]{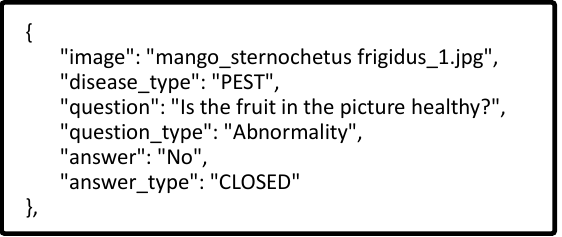}
    \caption{The JSON format of Agri-LLaVA-VQA-Bench.}
    \label{The JSON format of Agri-LLaVA-VQA-Bench.}
\end{figure*}

\section{Data Preprocessing}
\label{Data Preprocessing}
\begin{figure*}[h]
  \centering
  \includegraphics[width=0.45\textwidth]{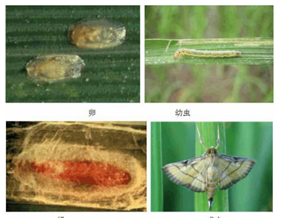}
  \hspace{1cm}
  \includegraphics[width=0.36\textwidth]{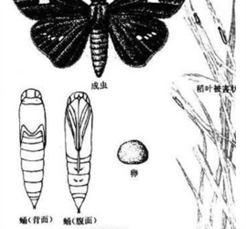}
  \caption{Examples of data preprocessing objects.}
  \label{Examples of data preprocessing objects}
\end{figure*}

For the downloaded dataset, we perform simple processing, mainly focusing on the IP102 dataset. On the one hand, we split images containing multiple sub-images to expand the dataset, on the other hand, we remove abstract images from it, as shown in Figure~\ref{Examples of data preprocessing objects}. For the processed images, to standardize the format, we uniformly name the images with the following format: \textit{crop category\_pest and disease name\_cardinal number.jpg}, for example: \textit{mango\_sternochetus frigidus\_1.jpg}.

\section{Data Cleaning}
\label{Data Cleaning}
Although we generate instruction data using instruction templates, the generated results are not satisfactory. Therefore, we clean the generated data to achieve the desired results.

While generating agricultural pests and diseases feature alignment data, we encountered issues with the designed format of two questions and two answers. Specifically, in the first question, we required the generated question to include the word ``image''. However, in actual generation results, this requirement was not consistently met. As a result, we made corrections to address this issue. Additionally, aside from the mentioned problem, unexpected outcomes occasionally arose when using GPT to generate ``answer''. GPT sometimes produced results based on its own judgment rather than adhering to the knowledge provided by the template. If this phenomenon is rare occurrence, we address it ourselves, on the contrary, if this phenomenon is widespread, we make improvements to the template. The final template is illustrated in Figure~\ref{prompt of stage-1 disease} and Figure~\ref{prompt of stage-1 pest}.

When generating the Agri-LLaVA-Chatbot-Bench, our purpose is to include a wide range of agricultural pest and disease questions in the generated conversations. We also aim to provide detailed answers for each type of question. However, during the generation process, we observed that some ``answer'' were not sufficiently detailed. To address this, we refined the generated ``answer''. Additionally, if there were an excessive number of questions related to the same category, such as symptoms, we removed some of them and designed other types of questions based on existing knowledge to ensure diversity in the questions.

\section{Data Supplement}
\label{Data Supplement}

\begin{table*}[h]
  \centering
	\renewcommand\arraystretch{1.2}
    \begin{tabular} {c|c|c}
    \toprule[1.5pt]
    \multicolumn{1}{p{8em}<{\centering}|}{Category} & \multicolumn{1}{p{20em}<{\centering}|}{Name}   & \multicolumn{1}{p{8em}<{\centering}}{Type} \\
    \midrule

    \multirow{3}[1]{*}{\fontsize{10}{14}\selectfont Strawberry} & {\fontsize{10}{14}\selectfont Strawberry ring spot disease} & Disease   \\
    \multicolumn{1}{c|}{} & {\fontsize{10}{14}\selectfont Strawberry skin rot } & Disease     \\
    \multicolumn{1}{c|}{} & {\fontsize{10}{14}\selectfont Strawberry bacterial wilt } & Disease     \\
    \midrule

    \multirow{4}[1]{*}{\fontsize{10}{14}\selectfont Apple} & {\fontsize{10}{14}\selectfont Apple yellow aphid} & Pest   \\
    \multicolumn{1}{c|}{} & {\fontsize{10}{14}\selectfont Apple blue flea Beetle} & Pest     \\
    \multicolumn{1}{c|}{} & {\fontsize{10}{14}\selectfont Apple mosaic } & Disease     \\
    \multicolumn{1}{c|}{} & {\fontsize{10}{14}\selectfont Apple powdery mildew } & Disease     \\
    \midrule

    \multirow{3}[1]{*}{\fontsize{10}{14}\selectfont Grape} & {\fontsize{10}{14}\selectfont Seudyra subflava Moore} & Pest   \\
    \multicolumn{1}{c|}{} & {\fontsize{10}{14}\selectfont Grape deadarm} & Disease     \\
    \multicolumn{1}{c|}{} & {\fontsize{10}{14}\selectfont Massonina viticola } & Disease     \\
    \midrule

    \multirow{3}[1]{*}{\fontsize{10}{14}\selectfont Banana} & {\fontsize{10}{14}\selectfont Banana cercospora leaf spot} & Disease   \\
   \multicolumn{1}{c|}{} & {\fontsize{10}{14}\selectfont Banana crown rot} & Disease     \\
    \multicolumn{1}{c|}{} & {\fontsize{10}{14}\selectfont Banana gray stripe disease } & Disease     \\
    \midrule
    
    \multirow{2}[1]{*}{\fontsize{10}{14}\selectfont Soybean} & {\fontsize{10}{14}\selectfont Soybean phytophthora root rot} & Disease   \\
    \multicolumn{1}{c|}{} & {\fontsize{10}{14}\selectfont Soybean fusarium wilt disease } & Disease     \\
    \midrule

    \multirow{2}[1]{*}{\fontsize{10}{14}\selectfont Peach} & {\fontsize{10}{14}\selectfont Peach brown rot} & Disease   \\
    \multicolumn{1}{c|}{} & {\fontsize{10}{14}\selectfont Peach fruit month} & Pest     \\
    \midrule

    \multirow{2}[1]{*}{\fontsize{10}{14}\selectfont Potato} & {\fontsize{10}{14}\selectfont Potato soft rot disease} & Disease   \\
    \multicolumn{1}{c|}{} & {\fontsize{10}{14}\selectfont Potato twenty-eight ladybird} & Pest     \\
    \midrule

    \multirow{2}[1]{*}{\fontsize{10}{14}\selectfont Sugarcane} & {\fontsize{10}{14}\selectfont Sugarcane pineapple disease} & Disease   \\
    \multicolumn{1}{c|}{} & {\fontsize{10}{14}\selectfont Sugarcane red rot disease} & Disease     \\
    \midrule

    \multirow{2}[1]{*}{\fontsize{10}{14}\selectfont Beet} & {\fontsize{10}{14}\selectfont Beet yellowing virus disease} & Disease   \\
    \multicolumn{1}{c|}{} & {\fontsize{10}{14}\selectfont Heterodera schachtii } & Pest     \\
    \midrule
    
    \multirow{2}[1]{*}{\fontsize{10}{14}\selectfont Citrus} & {\fontsize{10}{14}\selectfont Citrus melanose} & Disease   \\
    \multicolumn{1}{c|}{} & {\fontsize{10}{14}\selectfont Citrus sooty mold } & Disease     \\
    \midrule

    \multirow{1}[1]{*}{\fontsize{10}{14}\selectfont Blueberry} & {\fontsize{10}{14}\selectfont Blueberry gray mold} & Disease\\    \midrule
    
    \multirow{1}[1]{*}{\fontsize{10}{14}\selectfont Cassava} & {\fontsize{10}{14}\selectfont Cassava Mosaic} & Disease\\
    \midrule
 
    \multirow{1}[1]{*}{\fontsize{10}{14}\selectfont Tomato} & {\fontsize{10}{14}\selectfont Tomato yellow leaf curl virus disease} & Disease\\
    \midrule
 
    \multirow{1}[1]{*}{\fontsize{10}{14}\selectfont Corn} & {\fontsize{10}{14}\selectfont Corn stem rot} & Disease
    \\
    \bottomrule[1.5pt]
    
    \end{tabular}
  \vspace{1em}
  \caption{The components of Agri-LLaVA-Chatbot-Bench. ``Category'' indicates the crop species affected by pests and diseases, ``Name'' indicates the name of pests and diseases, and ``Type'' indicates whether it is a disease or pest.}
  \label{The components of Agri-LLaVA-Chatbot-Bench}
\end{table*}%

\begin{table*}[htbp]
  \centering
    \renewcommand\arraystretch{1.2}
    \begin{tabular}{p{3.9em}<{\centering}|p{11em}<{\centering}|p{2.5em}<{\centering}|p{12.5em}<{\centering}}
    \toprule[1.5pt]
    Category & Name  & \multicolumn{1}{c|}{Category} & \multicolumn{1}{p{15em}<{\centering}}{Name} \\
    \midrule
    
    \multirow{3}[2]{*}{Wheat}  & {\fontsize{9}{14}\ Wheat powdery mildew} & \multicolumn{1}{c|}{\multirow{3}[2]{*}{Wheat}} &  \multicolumn{1}{p{15em}<{\centering}}{\fontsize{9}{14}\ Wheat scab} \\
    \multicolumn{1}{c|}{}  & {\fontsize{9}{14}\ Wheat septoria} & & \multicolumn{1}{p{15em}<{\centering}}{\fontsize{9}{14}\ Wheat stem rust} \\
    \multicolumn{1}{c|}{} &{\fontsize{9}{14}\ Wheat chillella leaf blight} & & \multicolumn{1}{p{15em}<{\centering}}{\fontsize{9}{14}\ Wheat spindle streak mosaic disease} \\
    \midrule

    \multirow{5}[2]{*}{Rice}  & {\fontsize{9}{14}\ Rice bacterial streak spot} & \multicolumn{1}{c|}{\multirow{5}[2]{*}{Apple}} &  \multicolumn{1}{p{15em}<{\centering}}{\fontsize{9}{14}\ Apple alternaria leaf spot} \\
    \multicolumn{1}{c|}{}  & {\fontsize{9}{14}\ Rice flax spot} & & \multicolumn{1}{p{15em}<{\centering}}{\fontsize{9}{14}\ Apple grey spot} \\
    \multicolumn{1}{c|}{} &{\fontsize{9}{14}\ Leaf smut} & & \multicolumn{1}{p{15em}<{\centering}}{\fontsize{9}{14}\ Apple brown spot} \\
    \multicolumn{1}{c|}{}  & {\fontsize{9}{14}\ Rice koji disease} & & \multicolumn{1}{p{15em}<{\centering}}{\fontsize{9}{14}\ Apple powdery mildew} \\
    \multicolumn{1}{c|}{} &{\fontsize{9}{14}\ Rice sheath blight} & & \multicolumn{1}{p{15em}<{\centering}}{\fontsize{9}{14}\ Apple mosaic} \\
    \midrule
    
    \multirow{4}[2]{*}{Grape}  & {\fontsize{9}{14}\ Grape mosaic virus disease} & \multicolumn{1}{c|}{\multirow{4}[2]{*}{Tomato}} &  \multicolumn{1}{p{15em}<{\centering}}{\fontsize{9}{14}\ Tomato canker} \\
    \multicolumn{1}{c|}{}  & {\fontsize{9}{14}\ Grape downy mildew} & & \multicolumn{1}{p{15em}<{\centering}}{\fontsize{9}{14}\ Tomato mosaic virus} \\
    \multicolumn{1}{c|}{} &{\fontsize{9}{14}\ Grape powdery mildew} & & \multicolumn{1}{p{15em}<{\centering}}{\fontsize{9}{14}\ Tomato verticillium wilt} \\
    \multicolumn{1}{c|}{}  & {\fontsize{9}{14}\ Rhizopus stolnifer} & & \multicolumn{1}{p{15em}<{\centering}}{\fontsize{9}{14}\ Tomato yellow leaf curl virus} \\
    \midrule
    
    \multirow{4}[2]{*}{Tea}  & {\fontsize{9}{14}\ Tea algae leaf spot} & \multicolumn{1}{c|}{\multirow{4}[2]{*}{Cucumber}} &  \multicolumn{1}{p{15em}<{\centering}}{\fontsize{9}{14}\ Cucumber target spot} \\
    \multicolumn{1}{c|}{}  & {\fontsize{9}{14}\ Brown blight} & & \multicolumn{1}{p{15em}<{\centering}}{\fontsize{9}{14}\ Cucumber powdery mildew} \\
    \multicolumn{1}{c|}{} &{\fontsize{9}{14}\ Tea red leaf spot} & & \multicolumn{1}{p{15em}<{\centering}}{\fontsize{9}{14}\ Cucumber downy mildew} \\
    \multicolumn{1}{c|}{}  & {\fontsize{9}{14}\ Tea bird eye spot} & & \multicolumn{1}{p{15em}<{\centering}}{\fontsize{9}{14}\ Cucumber anthracnose} \\
    \midrule

    \multirow{3}[2]{*}{Pepper}  & {\fontsize{9}{14}\ Pepper virus disease} & \multicolumn{1}{c|}{\multirow{3}[2]{*}{Soybean}} &  \multicolumn{1}{p{15em}<{\centering}}{\fontsize{9}{14}\ Soybean root rot} \\
    \multicolumn{1}{c|}{}  & {\fontsize{9}{14}\ Pepper root rot} & & \multicolumn{1}{p{15em}<{\centering}}{\fontsize{9}{14}\ Soybean mosaic disease} \\
    \multicolumn{1}{c|}{}  & {\fontsize{9}{14}\ Pepper blossom end rot} & & \multicolumn{1}{p{15em}<{\centering}}{\fontsize{9}{14}\ Soybean bacterial spotted disease} \\
    \midrule
    
    \multirow{2}[1]{*}{Cowpea}  & {\fontsize{9}{14}\ Cowpea brown spot} & \multicolumn{1}{c|}{\multirow{2}[1]{*}{Potato}} &  \multicolumn{1}{p{15em}<{\centering}}{\fontsize{9}{14}\ Potato early blight} \\
    \multicolumn{1}{c|}{}  & {\fontsize{9}{14}\ Cowpea rust} & & \multicolumn{1}{p{15em}<{\centering}}{\fontsize{9}{14}\ Potato tuber hollow disease} \\
    \midrule
    
    \multirow{2}[1]{*}{Cashew}  & {\fontsize{9}{14}\ Gumosis} & \multicolumn{1}{c|}{\multirow{2}[1]{*}{Corn}} &  \multicolumn{1}{p{15em}<{\centering}}{\fontsize{9}{14}\ Corn spot} \\
    \multicolumn{1}{c|}{}  & {\fontsize{9}{14}\ Cashew anthracnose} & & \multicolumn{1}{p{15em}<{\centering}}{\fontsize{9}{14}\ Corn southern leaf blight} \\
    \midrule

    \multirow{1}[1]{*}{Beet}  & {\fontsize{9}{14}\ Cercospora leaf spots} & \multicolumn{1}{c|}{\multirow{1}[1]{*}{Lemon}} &  \multicolumn{1}{p{15em}<{\centering}}{\fontsize{9}{14}\ Lemon canker} \\
    \midrule
    
    \multirow{1}[1]{*}{Ash gourd}  & {\fontsize{9}{14}\ Potassium deficiency} & \multicolumn{1}{c|}{\multirow{1}[1]{*}{Bitter gourd}} &  \multicolumn{1}{p{15em}<{\centering}}{\fontsize{9}{14}\ Potassium deficiency} \\
   
    \bottomrule[1.5pt]
    \end{tabular}
  \vspace{1em}
  \caption{The components of the diseases in the Agri-LLaVA-VQA-Bench. ``Category'' indicates the crop species affected by diseases, ``Name'' indicates the name of diseases.}
  \label{The components of the diseases in the Agri-LLaVA-VQA-Bench}%
\end{table*}%

\begin{table*}[htbp]
  \centering

    \renewcommand\arraystretch{1.2}
    \begin{tabular}{p{4.5em}<{\centering}|p{12em}<{\centering}|c|c}
  
    \toprule[1.5pt]
    Category & Name  & \multicolumn{1}{p{3.5em}<{\centering}|}{Category} & \multicolumn{1}{p{14.7em}<{\centering}}{Name} \\
    \midrule
    
    \multirow{9}[2]{*}{Corn}  & {\fontsize{9}{14}\ Amsacta lactinea} & \multicolumn{1}{c|}{\multirow{9}[2]{*}{Citrus}} & \multicolumn{1}{p{14.7em}<{\centering}}{\fontsize{9}{14}\ Tetradacus c Bactrocera} \\
    \multicolumn{1}{c|}{}  & {\fontsize{9}{14}\ Spodoptera exigua Huner} & & \multicolumn{1}{p{14.7em}<{\centering}}{\fontsize{9}{14}\ Prodenia litura} \\
    \multicolumn{1}{c|}{} &{\fontsize{9}{14}\ Mythimnaseparata walker} & & \multicolumn{1}{p{14.7em}<{\centering}}{\fontsize{9}{14}\ Phyllocnistis citrella Stainton} \\
    \multicolumn{1}{c|}{}  & {\fontsize{9}{14}\ Fall armyworm} & & \multicolumn{1}{p{14.7em}<{\centering}}{\fontsize{9}{14}\ Toxoptera citricidus} \\
    \multicolumn{1}{c|}{} &{\fontsize{9}{14}\ Grass hoper} & & \multicolumn{1}{p{14.7em}<{\centering}}{\fontsize{9}{14}\ Parlatoria zizyphus Lucus} \\
    \multicolumn{1}{c|}{}  & {\fontsize{9}{14}\ Leaf beetle} & & \multicolumn{1}{p{14.7em}<{\centering}}{\fontsize{9}{14}\ Nipaecoccus vastalor} \\
    \multicolumn{1}{c|}{} &{\fontsize{9}{14}\ grub} & & \multicolumn{1}{p{14.7em}<{\centering}}{\fontsize{9}{14}\ Phyllocoptes oleiverus ashmead} \\
    \multicolumn{1}{c|}{}  & {\fontsize{9}{14}\ mole cricket} & & \multicolumn{1}{p{14.7em}<{\centering}}{\fontsize{9}{14}\ Toxoptera aurantii} \\
    \multicolumn{1}{c|}{} &{\fontsize{9}{14}\ Potosiabre vitarsis} & & \multicolumn{1}{p{14.7em}<{\centering}}{\fontsize{9}{14}\ } \\
    \midrule

    \multirow{7}[2]{*}{Mango}  & {\fontsize{9}{14}\ Dasineura sp} & \multicolumn{1}{c|}{\multirow{7}[2]{*}{Vitis}} & \multicolumn{1}{p{14.7em}<{\centering}}{\fontsize{9}{14}\ Apolygus lucorum} \\
    \multicolumn{1}{c|}{}  & {\fontsize{9}{14}\ Chlumetia transversa} & & \multicolumn{1}{p{14.7em}<{\centering}}{\fontsize{9}{14}\ Pseudococcus comstocki Kuwana} \\
    \multicolumn{1}{c|}{} &{\fontsize{9}{14}\ Sternochetus frigidus} & & \multicolumn{1}{p{14.7em}<{\centering}}{\fontsize{9}{14}\ Erythroneura apicalis} \\
    \multicolumn{1}{c|}{}  & {\fontsize{9}{14}\ Cicadellidae} & & \multicolumn{1}{p{14.7em}<{\centering}}{\fontsize{9}{14}\ parathrene regalis} \\
    \multicolumn{1}{c|}{} &{\fontsize{9}{14}\ Mango flat beak} & & \multicolumn{1}{p{14.7em}<{\centering}}{\fontsize{9}{14}\ Polyphagotars onemus latus} \\
    \multicolumn{1}{c|}{}  & {\fontsize{9}{14}\ Deporaus marginatus Pascoe} & & \multicolumn{1}{p{14.7em}<{\centering}}{\fontsize{9}{14}\ Brevipoalpus lewisi McGregor} \\
    \multicolumn{1}{c|}{} &{\fontsize{9}{14}\ } & & \multicolumn{1}{p{14.7em}<{\centering}}{\fontsize{9}{14}\ Colomerus vitis} \\
    \midrule

    \multirow{5}[2]{*}{Rice}  & {\fontsize{9}{14}\ white backed plant} & \multicolumn{1}{c|}{\multirow{5}[2]{*}{Beet}} & \multicolumn{1}{p{14.7em}<{\centering}}{\fontsize{9}{14}\ cabbage army worm} \\
    \multicolumn{1}{c|}{}  & {\fontsize{9}{14}\ Hispa} & & \multicolumn{1}{p{14.7em}<{\centering}}{\fontsize{9}{14}\ sericaorient alismots chulsky} \\
    \multicolumn{1}{c|}{} &{\fontsize{9}{14}\ Rice Stemfly} & & \multicolumn{1}{p{14.7em}<{\centering}}{\fontsize{9}{14}\ Beet spot flies} \\
    \multicolumn{1}{c|}{}  & {\fontsize{9}{14}\ paddy stem maggot} & & \multicolumn{1}{p{14.7em}<{\centering}}{\fontsize{9}{14}\ beet army worm} \\
    \multicolumn{1}{c|}{} &{\fontsize{9}{14}\ grain spreader thrips} & & \multicolumn{1}{p{14.7em}<{\centering}}{\fontsize{9}{14}\ beet fly} \\
    \midrule

    \multirow{3}[2]{*}{Wheat}  & {\fontsize{9}{14}\ english grain aphid} & \multicolumn{1}{c|}{\multirow{3}[2]{*}{Wheat}} & \multicolumn{1}{p{14.7em}<{\centering}}{\fontsize{9}{14}\ longlegged spider mite} \\
    \multicolumn{1}{c|}{}  & {\fontsize{9}{14}\ cerodonta denticornis} & & \multicolumn{1}{p{14.7em}<{\centering}}{\fontsize{9}{14}\ wheat phloeothrips} \\
    \multicolumn{1}{c|}{} &{\fontsize{9}{14}\ penthaleus major} & & \multicolumn{1}{p{14.7em}<{\centering}}{\fontsize{9}{14}\ green bug} \\
    \midrule

    \multirow{1}[2]{*}{Cabbage}  & {\fontsize{9}{14}\ Looper} & \multicolumn{1}{c|}{\multirow{1}[2]{*}{Tomato}} & \multicolumn{1}{p{14.7em}<{\centering}}{\fontsize{9}{14}\ Leaf miner} \\
    \midrule
    
    \multirow{1}[2]{*}{Alfalfa}  & {\fontsize{9}{14}\ alfalfa seed chalcid} & \multicolumn{1}{c|}{\multirow{1}[2]{*}{Alfalfa}} & \multicolumn{1}{p{14.7em}<{\centering}}{\fontsize{9}{14}\ odontothrips loti} \\

    \bottomrule[1.5pt]
    \end{tabular}%
  \vspace{1em}
  \caption{The components of the pests in the Agri-LLaVA-VQA-Bench. ``Category'' indicates the crop species affected by pests, ``Name'' indicates the name of pests.}
  \label{The components of the pests in the Agri-LLaVA-VQA-Bench}%
\end{table*}%

\begin{figure*}
    \centering \includegraphics[width=0.53\linewidth]{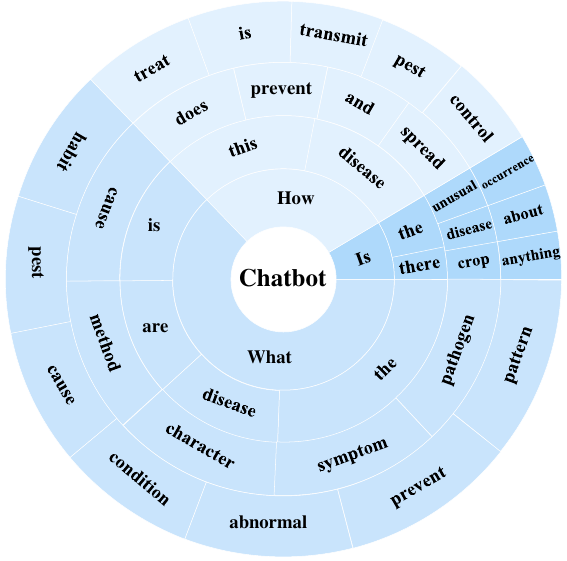}
    \caption{The top 4 words of the top 3 questions in the Agri-LLaVA-Chatbot-Bench.}
    \label{The top 4 words of the top 3 questions in the Agri-LLaVA-Chatbot-Bench.}
\end{figure*}

Figure~\ref{The top 4 words of the top 3 questions in the Agri-LLaVA-Chatbot-Bench.} represents the Agri-LLaVA-Chatbot-Bench, displaying the top four most frequently occurring words in top three questions, arranged from inside to outside. It is divided into three regions, with each region corresponding to questions starting with the respective word, for example: ``What''.

\begin{figure*}[h]
    \centering
    \includegraphics[width=0.7\linewidth]{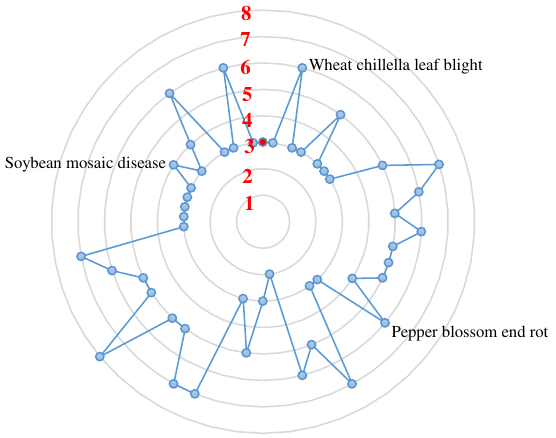}
    \caption{The distribution of the number of different types of diseases in the Agri-LLaVA-VQA-Bench.}
    \label{The distribution of the number of different types of diseases in the Agri-LLaVA-VQA-Bench.}
\end{figure*}

In addition to showing the composition of the Agri-LLaVA-VQA-Bench, we also counted the number distribution of each pests and diseases, as shown in Figure~\ref{The distribution of the number of different types of diseases in the Agri-LLaVA-VQA-Bench.}.

Figure~\ref{The distribution of the number of different types of diseases in the Agri-LLaVA-VQA-Bench.} consists of eight concentric circles, representing the number of image of diseases from 1 to 8, and each dot in the figure above represents a disease, there are 49 diseases in total. The red-marked points in the figure represent the first disease, which is ``Wheat powdery mildew'' in Table~\ref{The components of the diseases in the Agri-LLaVA-VQA-Bench}. Moving clockwise, they correspond to diseases listed in the ``Name'' column of Table~\ref{The components of the diseases in the Agri-LLaVA-VQA-Bench} from the top-left to the bottom-left, and from the top-right to the bottom-right. The figure annotates three diseases as reference examples, indicating the relationship between the corresponding diseases and their quantities.

\begin{figure*}
    \centering
    \includegraphics[width=0.85\linewidth]{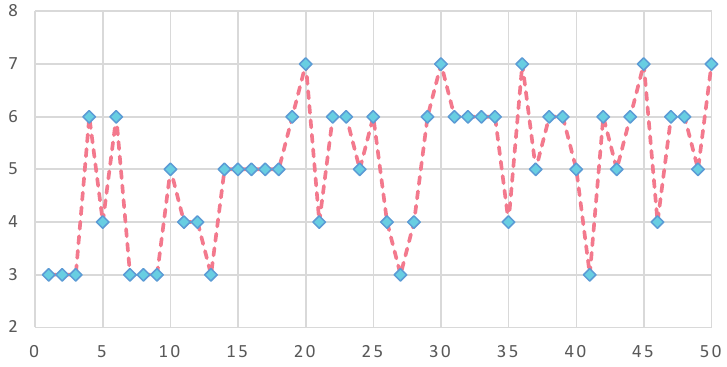}
    \caption{The distribution of the number of different types of pests in the Agri-LLaVA-VQA-Bench.}
    \label{The distribution of the number of different types of pests in the Agri-LLaVA-VQA-Bench.}
\end{figure*}

In Figure~\ref{The distribution of the number of different types of pests in the Agri-LLaVA-VQA-Bench.}, the horizontal axis ranges from 1 to 50, representing the 50 types of pests in Table~\ref{The components of the pests in the Agri-LLaVA-VQA-Bench}. Specifically, 1 to 25 represent the top-down 25 types of pests in the left column of Table~\ref{The components of the pests in the Agri-LLaVA-VQA-Bench}, while 26 to 50 represent the top-down 25 types of pests in the right column of Table~\ref{The components of the pests in the Agri-LLaVA-VQA-Bench}. The vertical axis represents the quantities of image of each type of pest in the Agri-LLaVA-VQA-Bench.
We conducted a statistical analysis on the questions in the Agri-LLaVA-VQA-Bench. Figure~\ref{The distribution of the number of different types of pests in the Agri-LLaVA-VQA-Bench.} shows the results.

\begin{figure*}
    \centering
    \includegraphics[width=0.85\linewidth]{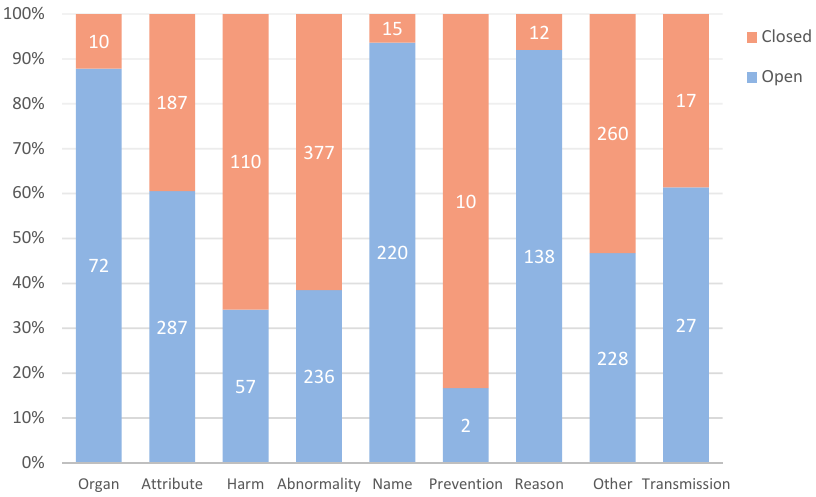}
    \caption{Distribution of open-set and closed-set in 9 types of questions in Agri-LLaVA-VQA-Bench.}
    \label{Distribution of open-set and closed-set in 9 types of questions in Agri-LLaVA-VQA-Bench.}
\end{figure*}

Figure~\ref{Distribution of open-set and closed-set in 9 types of questions in Agri-LLaVA-VQA-Bench.} illustrates the categories of questions in the Agri-LLaVA-VQA-Bench and their respective quantities, categorized into open-ended and closed-ended questions. The vertical axis represents the ratio, which can clearly show the proportion of open-set and closed-set in different kinds of question.

\section{More Results}
\label{More Results}

\begin{figure*}
    \centering
    \includegraphics[width=0.85\linewidth]{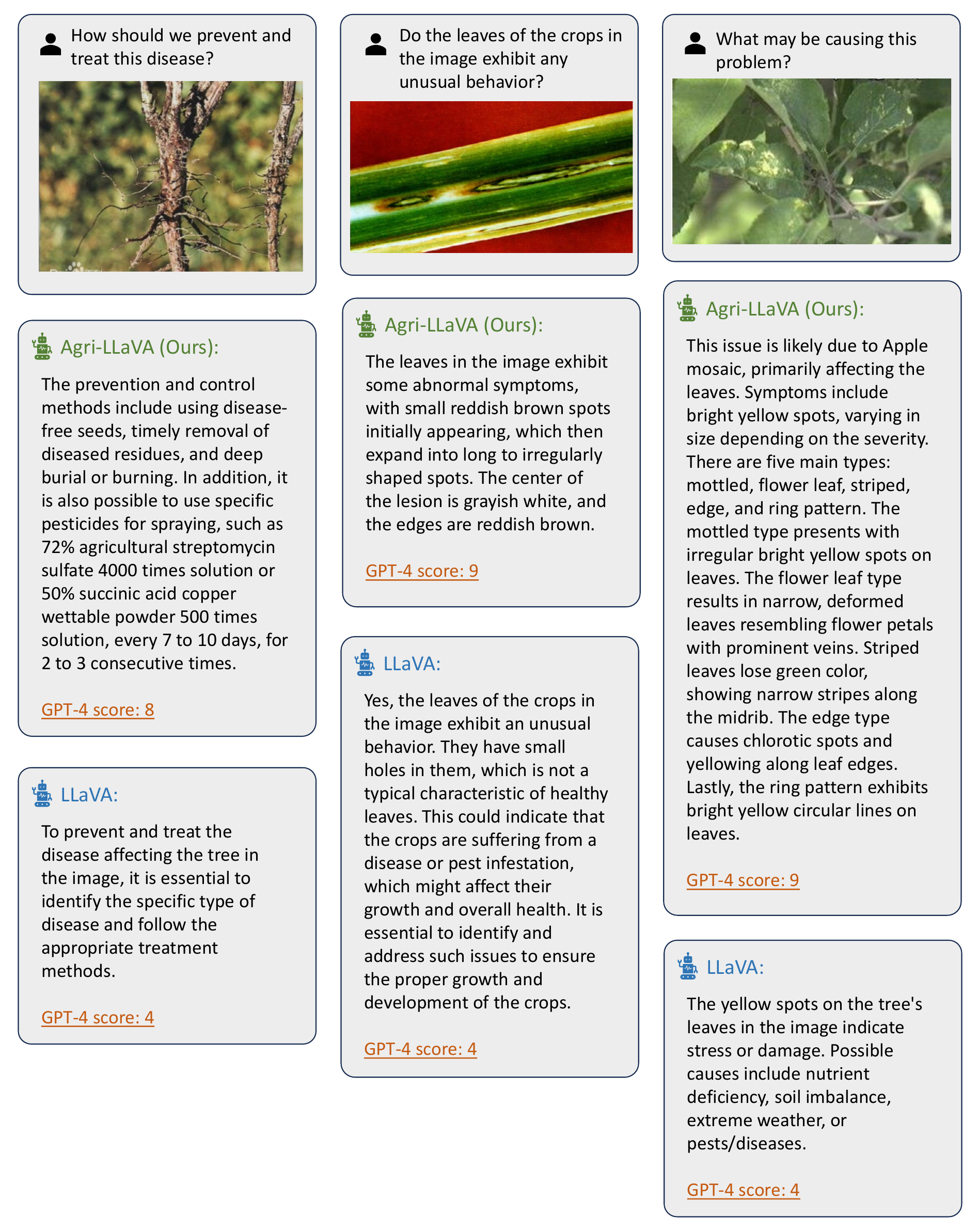}
    \caption{More visual examples of Agri-LLaVA on multimodal conversations.}
    \label{More visual examples of Agri-LLaVA on multimodal conversations}
\end{figure*}

Additional visualization results for a wider range of Agri-LLaVA are available in Figure~\ref{More visual examples of Agri-LLaVA on multimodal conversations}, providing further insights into model performance and comparisons across various scenarios.

\section{Limitations}
\label{Limitations}

\begin{figure*}
    \centering
    \includegraphics[width=0.85\linewidth]{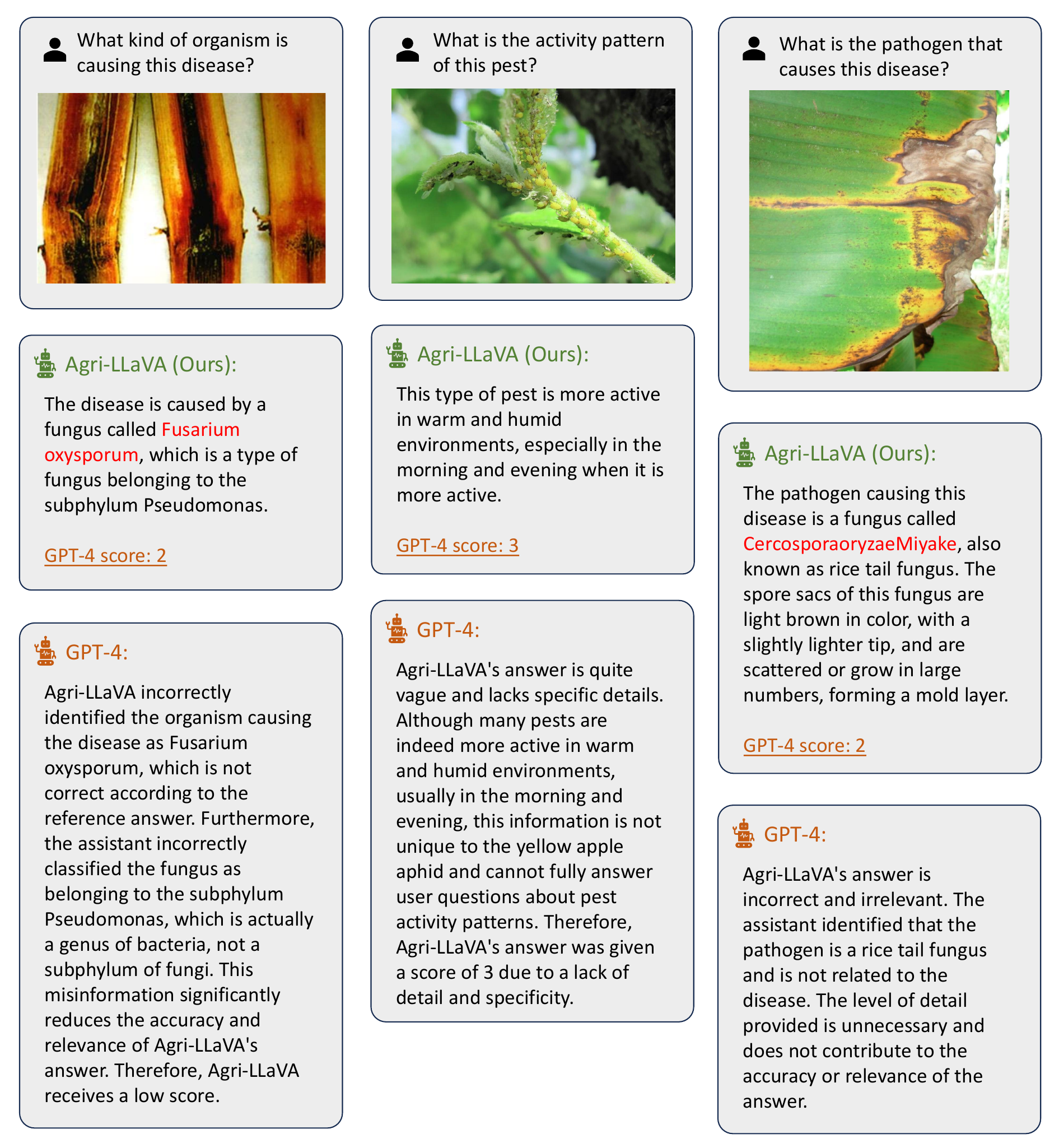}
    \caption{Some failure cases of Agri-LLaVA. Due to the complexity of agricultural pest and disease images, Agri-LLaVA can still misdiagnose certain conditions. This is particularly evident when inquiring about deep-level causes, such as specific information about pathogens. Such misdiagnoses may occur because features in the images resemble those of other diseases, or because the model lacks sufficient knowledge of specific details. This indicates that, despite extensive agricultural expertise injected during training, the model still requires further optimization and improvement to enhance its accuracy and diagnostic capabilities in complex agricultural environments.}
    \label{Some failure cases of Agri-LLaVA}
\end{figure*}

Despite substantial efforts in data collection, the scarcity of agricultural data means that certain pest and disease types remain underrepresented. Consequently, Agri-LLaVA may exhibit suboptimal performance in some agricultural scenarios. Figure~\ref{Some failure cases of Agri-LLaVA} illustrates several examples of these failure cases.

\section{Broader Impact}
\label{Broader Impact}

Agri-LLaVA, as the first open-source large multimodal model tailored for agriculture, holds tremendous potential for agricultural intelligence. 
However, it may also face several potential risks and unresolved issues. 
Some of these issues are similar to those encountered by general LMMs, but there are also unique challenges arising from the specific nature of agricultural scenarios. 
When considering the promotion and application of such models, we must carefully address these challenges.

One such challenge is \textbf{data privacy and security risks}. 
Agricultural multimodal assistants rely on extensive data, including field information, meteorological data, and crop growth data. 
However, these data involve the privacy of farmers and agricultural producers. 
Any leakage or misuse of this data could lead to serious privacy and security issues.

Another issue is \textbf{insufficient generalization capability}. 
Due to the scarcity and instability of agricultural data quality, initial models developed may lack the ability to generalize well to new datasets, thereby failing to adapt effectively to various agricultural environments and scenarios.

Moreover, \textbf{misleading predictions and decision-making risks} are also significant concerns. 
Although agricultural multimodal assistants can provide real-time predictions and decision support for agricultural pests and diseases, the complexity of agricultural ecosystems and the uncertainty of environmental factors may result in prediction errors and uncertainties, potentially leading to misleading decisions and losses.

Lastly, \textbf{technical dependencies and security vulnerabilities} also need attention. Agricultural multimodal assistants rely on advanced technologies and system support, such as machine learning algorithms, cloud computing platforms, and sensor technologies. 
If these technologies fail or become unstable, it may affect the accuracy and reliability of the models, thereby impacting agricultural production. Additionally, if the security of agricultural multimodal assistants is not robust enough, they may face the risk of being attacked or maliciously manipulated, resulting in losses to agricultural production and farmer interests.

Although the aforementioned issues may exist, Agri-LLaVA also brings more benefits than drawbacks to the community. 
Agri-LLaVA opens up new possibilities for agricultural LMMs, injecting fresh vitality into the technological development of the agricultural sector. 
Our endeavor lays the groundwork for future work, allowing various specific aspects of agricultural large models to organize data and train models following our process. 
The community can further research based on our model and take measures to mitigate and avoid potential risks. 
The open-source nature of Agri-LLaVA can stimulate the development of the field, fostering knowledge sharing and collaboration, thereby incubating new ideas and driving innovation and progress in agricultural technology.

\section{Evaluation Metrics}
\label{Evaluation Metrics}
To evaluate the performance of the model on the Agri-LLaVA-VQA-Bench, we use both F1-score and Accuracy as evaluation metrics for the open-set and closed-set portions of the Agri-LLaVA-VQA-Bench, respectively.

\textbf{F1-score.}
F1-score is the harmonic mean of precision and recall, calculated using the following formula:
\\
\begin{equation}
\label{eq}
 F_{1} = 2 \times \frac{Precision \times Recall}{Precision + Recall}, \\ 
\end{equation}
\\
The calculation formulas for precision and recall are as follows:
\\
\begin{equation}
\label{eq:Positional Encoding}
  Precision = {TP\over TP+FP},
\end{equation}

\begin{equation}
\label{eq}
 Recall = {TP\over TP+FN},
\end{equation}

Among them, TP represents the number of true positives, meaning the number of positive samples correctly identified by the model; FP represents the number of false positives, indicating the number of negative samples incorrectly identified as positive; FN represents the number of false negatives, denoting the number of positive samples missed by the model.

\textbf{Accuracy.}
Accuracy represents the proportion of correctly predicted samples, and its calculation formula is as follows:
\\
\begin{equation}
\label{eq}
 Accuracy = \frac{TP+TN}{TP+TN+FP+FN},
\end{equation}

Among them, FP represents the number of false positives, indicating the number of negative samples incorrectly identified as positive.

\end{document}